\documentclass[sigconf,nonacm]{acmart}
\usepackage{booktabs}
\usepackage{array}
\usepackage{multirow}
\usepackage{tikz}
\usetikzlibrary{arrows.meta,positioning,fit,calc,backgrounds,shapes.geometric,decorations.pathreplacing,patterns}
\usepackage{xcolor}
\usepackage{pifont}
\usepackage{colortbl}
\definecolor{oursbg}{RGB}{254,237,222}
\definecolor{acc}{RGB}{38,113,206}  %
\usepackage{algorithm}
\usepackage{algpseudocode}
\newcommand{\drname}{Needwise}  %
\newcommand{\dr}{\textsc{\drname}}
\newcommand{\drh}{\textsc{\drname-H}}
\newcommand{\drf}{\textsc{\drname-F}}
\newcommand{\drhf}{\textsc{\drname-HF}}

\newcommand{\supp}{\textsf{supported}}
\newcommand{\contra}{\textsf{contradicted}}
\newcommand{\unk}{\textsf{unknown}}

\renewcommand\footnotetextcopyrightpermission[1]{}

\begin{document}

\title{\drname: Generation-Free Conversational Recommendation from Need-Level Typed Decisions}

\author{Sungchul Kim}
\affiliation{\institution{Adobe Research}\city{San Jose}\country{USA}}
\email{sukim@adobe.com}
\author{Weixiao Huang}
\affiliation{\institution{Adobe Inc.}\city{San Jose}\country{USA}}
\author{Mekala Raja Sekhar Reddy}
\affiliation{\institution{Adobe Inc.}\city{San Jose}\country{USA}}
\author{Amrutha Srinivasan}
\affiliation{\institution{Adobe Inc.}\city{San Jose}\country{USA}}

\begin{abstract}
Conversational recommenders must rank items against needs that catalogs record only in part, while users add,
withdraw and reverse needs and switch between requests. LLM rankers re-read the dialogue at every turn: they are
slow, and their ordering does not show which needs an item meets. We show that a conversational recommender can be
built from typed decisions instead: named questions over fixed options, answered with option probabilities rather
than generated text. On each turn, one typed request maps the message onto the catalog's fields, keeps what no field
expresses in the user's words, and decides how the turn changes each earlier need. \dr{} then ranks candidates
must-first: the needs on the kind of item, its place and its price form a tier, and within it one holistic decision
and need-level verdicts (supported, contradicted or unknown) are combined with equal weight (\drhf{}). With the graded attributes withheld from the records, so that needs must be inferred from names,
descriptions and reviews, \drhf{} ranks Google Local requests significantly better than GPT-6 Sol
(0.383 against 0.343 nDCG@10) and is not significantly different from the strongest
LLM ranker on Google Local and Yelp; these comparisons keep their sign under every treatment of unrecorded attributes
we test. It costs a fraction of GPT-6 Sol's API price and answers in 2.5\,s at the median on Google
Local, against 5.6\,s for GPT-6 Sol. In conversations, the typed state keeps the ranking across switches and returns, but
it recognizes only 19--27\% of withdrawn needs.
\end{abstract}

\keywords{conversational recommendation, typed decision models, query-time features, preference state tracking, evidence-based ranking}

\maketitle

\section{Introduction}
\label{sec:intro}
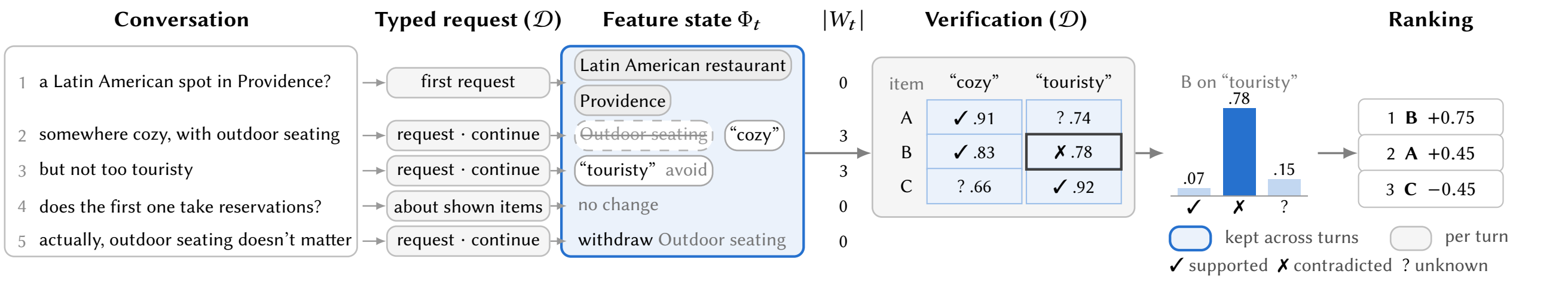
\begin{figure*}[t]
\centering
\begin{tikzpicture}[font=\sffamily,
  lab/.style={font=\sffamily\scriptsize, text=black!55, inner sep=1pt, align=center},
  ttl/.style={font=\sffamily\footnotesize\bfseries, inner sep=1pt, anchor=base},
  gbox/.style={draw=black!25, fill=black!4, rounded corners=3pt, line width=0.5pt},
  wbox/.style={draw=black!25, fill=white, rounded corners=3pt, line width=0.5pt},
  abox/.style={draw=acc, fill=acc!9, rounded corners=3pt, line width=0.8pt},
  arr/.style={-{Latex[length=1.6mm,width=1.3mm]}, line width=0.65pt, draw=black!50},
  sarr/.style={-{Latex[length=1.1mm,width=0.9mm]}, line width=0.45pt, draw=black!40},
  msg/.style={font=\sffamily\scriptsize, anchor=west, inner sep=1pt},
  act/.style={gbox, font=\sffamily\scriptsize, inner sep=1.2pt, minimum width=18mm, minimum height=3.3mm},
  chip/.style={rounded corners=4pt, font=\sffamily\scriptsize, inner sep=1.5pt, minimum height=3.3mm, anchor=west,
    line width=0.5pt},
  pred/.style={chip, draw=black!28, fill=black!7},
  feat/.style={chip, draw=black!35, fill=white},
  gone/.style={chip, draw=black!30, dashed, text=black!40, fill=white},
  cell/.style={draw=acc!45, fill=acc!9, minimum width=10.5mm, minimum height=3.9mm, font=\sffamily\scriptsize,
    inner sep=0pt},
  rk/.style={draw=black!25, fill=white, rounded corners=2pt, minimum width=16mm, minimum height=4.1mm,
    font=\sffamily\scriptsize, inner sep=1pt}]
\def\ra{2.755}\def\rb{2.17}\def\rc{1.78}\def\rd{1.39}\def\re{1.00}
\def\fy{1.975}%
\node[ttl] at (1.98,3.38) {Conversation};
\node[ttl] at (5.15,3.38) {Typed request ($\mathcal{D}$)};
\node[ttl] at (7.52,3.38) {Feature state $\Phi_t$};
\node[ttl] at (9.30,3.38) {$|W_t|$};
\node[ttl] at (11.10,3.38) {Verification ($\mathcal{D}$)};
\node[ttl] at (15.80,3.38) {Ranking};
\foreach \n/\y/\t/\a in {1/\ra/{a Latin American spot in Providence?}/{first request},
    2/\rb/{somewhere cozy, with outdoor seating}/{request $\cdot$ continue},
    3/\rc/{but not too touristy}/{request $\cdot$ continue},
    4/\rd/{does the first one take reservations?}/{about shown items},
    5/\re/{actually, outdoor seating doesn't matter}/{request $\cdot$ continue}} {
  \node[msg, text=black!45] at (0.12,\y) {\n};
  \node[msg] at (0.36,\y) {\t};
  \node[act] (a\n) at (5.15,\y) {\a};
  \draw[sarr] (3.98,\y) -- (a\n.west);
}
\begin{scope}[on background layer]
\node[wbox, fit={(0.02,3.16) (3.92,0.83)}, inner sep=0pt] {};
\end{scope}
\node[pred] (s1) at (6.32,2.95) {Latin American restaurant};
\node[pred] (s2) at (6.32,2.56) {Providence};
\node[gone] (s3) at (6.32,\rb) {Outdoor seating};
\draw[black!45, line width=0.5pt] ([xshift=2pt]s3.west) -- ([xshift=-2pt]s3.east);
\node[feat] (s4) at ([xshift=1.2mm]s3.east) {``cozy''};
\node[feat] (s5) at (6.32,\rc) {``touristy''\enspace\textcolor{black!50}{avoid}};
\node[lab, anchor=west] (s6) at (6.32,\rd) {no change};
\node[lab, anchor=west, text=black] (s7) at (6.32,\re) {withdraw \textcolor{black!55}{Outdoor seating}};
\foreach \n/\y in {1/\ra, 2/\rb, 3/\rc, 4/\rd, 5/\re} {\draw[sarr] (a\n.east) -- (6.24,\y);}
\begin{scope}[on background layer]
\node[abox, fit={(6.18,3.16) (8.86,0.83)}, inner sep=0pt] (state) {};
\end{scope}
\foreach \y/\w in {\ra/0, \rb/3, \rc/3, \rd/0, \re/0} {\node[font=\sffamily\scriptsize, inner sep=1pt] at (9.30,\y) {\w};}
\node[lab] (hz) at (10.00,2.75) {item};
\node[lab, text=black] (h1) at (10.75,2.75) {``cozy''};
\node[lab, text=black] (h2) at (11.85,2.75) {``touristy''};
\foreach \r/\y/\a/\b in {A/2.37/\ding{51}\ .91/?\ .74, B/1.99/\ding{51}\ .83/\ding{55}\ .78, C/1.61/?\ .66/\ding{51}\ .92} {
  \node[lab, text=black] at (10.00,\y) {\r};
  \node[cell] (x\r1) at (10.75,\y) {\a};
  \node[cell] (x\r2) at (11.85,\y) {\b};
}
\node[draw=black!75, line width=0.8pt, minimum width=10.5mm, minimum height=3.9mm, inner sep=0pt] at (11.85,1.99) {};
\begin{scope}[on background layer]
\node[gbox, fit=(hz)(h2)(xA2)(xC1)(xC2), inner sep=4pt] (dec) {};
\end{scope}
\def\bx{13.00}\def\by{1.50}\def\bh{1.25}
\node[lab] at (\bx+0.68,2.75) {B on ``touristy''};
\foreach \i/\p/\s/\c in {0/.07/\ding{51}/acc!28, 1/.78/\ding{55}/acc, 2/.15/?/acc!28} {
  \fill[\c] (\bx+\i*0.5,\by) rectangle ++(0.36,\p*\bh);
  \node[font=\sffamily\scriptsize, inner sep=1pt, anchor=south] at (\bx+\i*0.5+0.18,\by+\p*\bh) {\p};
  \node[font=\sffamily\scriptsize, inner sep=1pt, anchor=north] at (\bx+\i*0.5+0.18,\by) {\s};
}
\draw[black!40, line width=0.4pt] (\bx-0.08,\by) -- (\bx+1.44,\by);
\node[rk] (r1) at (15.80,2.37) {1\enspace\textbf{B}\enspace $+0.75$};
\node[rk] (r2) at (15.80,1.975) {2\enspace\textbf{A}\enspace $+0.45$};
\node[rk] (r3) at (15.80,1.58) {3\enspace\textbf{C}\enspace $-0.45$};
\draw[arr] (state.east |- 0,\fy) -- (dec.west |- 0,\fy);
\draw[arr] (dec.east |- 0,\fy) -- (\bx-0.15,\fy);
\draw[arr] (\bx+1.55,\fy) -- (r2.west);
\node[abox, minimum width=4.5mm, minimum height=2.6mm, inner sep=0pt, anchor=west] (lg1) at (12.90,1.03) {};
\node[lab, anchor=west, text=black!70] (lt1) at ([xshift=1mm]lg1.east) {kept across turns};
\node[gbox, minimum width=4.5mm, minimum height=2.6mm, inner sep=0pt, anchor=west] (lg2) at ([xshift=3mm]lt1.east) {};
\node[lab, anchor=west, text=black!70] at ([xshift=1mm]lg2.east) {per turn};
\node[lab, anchor=west, text=black!70] at (12.87,0.72) {\textcolor{black}{\ding{51}} supported\enspace
  \textcolor{black}{\ding{55}} contradicted\enspace \textcolor{black}{?} unknown};
\end{tikzpicture}
\caption{Overview of \dr{}. Per turn, a typed request predicts whether the message is a request, which field values
(grey) and needs in the user's words (quoted) it adds, and what it changes. Only undecided pairs are judged ($|W_t|$;
blue: stored, reused); \drf{} averages the signed evidence $p_s-p_c$ over the needs. Turn predictions are real; other values illustrative.}
\label{fig:overview}
\end{figure*}

A conversational recommender takes needs that a user states in natural language and ranks the items of a catalog by
how well they meet them~\cite{christakopoulou2016towards,jannach2021survey}. Two properties of real requests make this
hard. First, needs exceed the catalog schema. In ``a Latin American spot in Providence, somewhere cozy with outdoor
seating, but not too touristy'', category, town and outdoor seating map to fields, whereas coziness and tourist appeal
appear, if at all, only in descriptions and reviews. Similarity does not establish satisfaction: a review complaining
about noise resembles a request for a quiet place. Second, needs change. Users withdraw needs, reverse them, switch to
another request and return to an earlier one. LLM-based recommenders read the dialogue and the candidates and rank
them~\cite{he2023llmcrs,hou2024llmrank}. However, they re-read everything at every turn, and their ordering shows
neither which needs a candidate meets nor whether an unmet need is contradicted or merely undocumented.

Typed decision models offer another primitive~\cite{typesafeJev2026,msDecision1,openaiDecisions2026,cloudflareClef}.
They take a state and named questions, each with a finite option set, and return a distribution over every question's
options in one forward pass, without generating text. Each answer is a typed, inspectable statement about one
candidate, many are answered in one request, and answers can be stored and reused. We ask how to build a conversational
recommender from such decisions alone, which requires deciding \emph{what} to ask, \emph{how} to read the message,
and \emph{when} to ask.

\emph{What to ask.} \dr{} asks one question per candidate and need, with three options---the record supports the
need, contradicts it, or gives insufficient evidence---and ranks by signed expected evidence, with hard schema
predicates first (\drf{}). The third option matters: a record silent on a property is no evidence that the item lacks
it. A holistic question per candidate (\drh{}) instead judges the request as a whole. \drhf{} combines the two
must-first: the hard predicates on the kind of item, place and price form a tier, and within it the holistic decision
and the need-level evidence are added with equal weight (Figure~\ref{fig:overview}). A must is never outweighed, and
nothing is tuned.

\emph{How to read the message.} Turning a message into needs is usually left to a generative LLM, which adds a second
kind of model to every turn and re-reads the dialogue each time~\cite{he2023llmcrs}. Interpretation can itself be
typed. \dr{} retrieves a shortlist of the catalog's field values for the message, and one typed request asks which of
them the message asks for, rules out or leaves alone. Whatever no field expresses is kept in the user's own words and
verified as such, and so is the kind of item, which catalog values state only coarsely. No text is generated at query
time.

\emph{When to ask.} The same typed request asks whether the turn is a request at all and how it changes each earlier
need. The request in force is a set of \emph{query-time features} with stable identities. The transition between
consecutive states (features added, withdrawn, reversed or restored) determines the decisions a turn requires: a
non-request turn retrieves, judges and re-ranks nothing, and withdrawn or reversed needs reuse stored verdicts.

We evaluate on Google Local and Yelp with their graded attributes withheld from the records, so needs must be
inferred from names, descriptions and reviews, against re-rankers, LLM rankers up to GPT-6 Sol, an agent and stateful
rewriting on the same pools. Grades treat an attribute the source does not record as absent; we bound every comparison
by resolving such attributes against and for each item. Without adapting model weights, \drhf{} ranks Google Local
requests significantly better than GPT-6 Sol and is not significantly different from the strongest LLM ranker on
Google Local and Yelp, at a fraction of GPT-6 Sol's API cost and with a lower median latency than the strongest LLM rankers. In conversations, a holistic decision that
reads the typed state instead of the dialogue keeps the ranking across switches and returns. Its main weakness is
withdrawals: the typed request recognizes only 19\% of them on Google Local and
27\% on Yelp.

Our contributions are:
\begin{itemize}
\item \textbf{Typed conversational recommendation:} one typed request maps each message onto the catalog's fields and
keeps the rest in the user's words, and candidates are ranked by need-level verdicts with an insufficient-evidence
option, a holistic decision, or both, must-first.
\item \textbf{Transition-routed decisions:} typed questions about the turn and each earlier need decide which judgments
a turn requires, so withdrawals and reversals ask nothing new.
\item \textbf{An empirical study} on two datasets of single-request ranking, state tracking in conversations,
latency and cost, with every comparison bounded under unrecorded attributes.
\end{itemize}

\section{Related Work}
\label{sec:related}

\paragraph{Typed decision models.} Typed interfaces return option probabilities for questions about a
state~\cite{typesafeJev2026,msDecision1,openaiDecisions2026,cloudflareClef}; open families add option-order consistency
training~\cite{dhankhar2026drex} and looped computation~\cite{berghaus2026reload}, and can replace LLM judges on many
benchmarks~\cite{subramanian2026jevornot}. Studies cover reranking~\cite{lyu2026decisionrerank}, efficiency~\cite{tang2026audit}, option and label
effects~\cite{rajput2026iiajev,sun2026optionname,azizi2026labeloverride}, wording and order
sensitivity~\cite{sharma2026typedbench,pai2026jevbench,imperatrice2026orderbias} and rejection~\cite{zhong2026rejectionbottleneck}. DecisionRank distills a listwise teacher into a decision-model
reranker~\cite{wu2026decisionrank}, others use calibration objectives~\cite{gao2026openjevrlcd,deka2026sbert2s1}, and
decision fine-tuning has not beaten reranker fine-tuning of the same backbone~\cite{dinzinger2026systemone}. We instead
build every query-time judgment---reading the message, tracking the state and verifying candidates---from typed
decisions, without adapting the model.

\paragraph{LLM ranking and verification.} LLM rankers use pointwise label probabilities~\cite{zhuang2024finegrained},
listwise comparisons~\cite{sun2023rankgpt} and reasoning~\cite{weller2025rank1}; multi-condition
ranking~\cite{pezeshkpour2024mcrank} and negation~\cite{weller2024nevir} remain difficult. Decomposing instructions into
checks~\cite{qin2024infobench,cook2024tick} motivates our per-feature evaluation, and the three-way relation follows fact
verification~\cite{thorne2018fever}. Probabilities can come from option tokens or verbal
reports~\cite{zhuang2024finegrained,tian2023justask}.

\paragraph{Conversational recommendation and state tracking.} Attribute-based conversational recommenders elicit and
revise preferences over a predefined vocabulary~\cite{sun2018conversational,zhang2018towards,jannach2021survey}, and
critiquing refines item features in natural language~\cite{wu2019critiquing}. LLMs support zero-shot
recommendation~\cite{he2023llmcrs,hou2024llmrank}, tool-using agents~\cite{huang2023interecagent,zhao2024toolrec} and
simulated-user evaluation~\cite{wang2023ievalm}; reinforcement learning adapts LLMs that generate recommendation lists or
queries for a fixed recommender~\cite{zhu2025rankgrpo,lin2025recr1}. Conversational search keeps context through query
rewriting~\cite{elgohary2019canard,mao2023llm4cs}, including rewriters trained with retrieval
rewards~\cite{zhu2025convsearchr1}, or revisable memory~\cite{ali2026memrevise}, and schema-guided dialogue state
tracking predicts slot values from natural-language slot descriptions~\cite{rastogi2020sgd}. Typed interpretation tracks
state in this spirit, with a decision model over the catalog's own fields, but keeps what no field expresses in the
user's words, links every need to candidate-level evidence and decides from its transitions which judgments a turn
requires.

\paragraph{Schema-based retrieval and attribute inference.} Self-query retrieval maps requests to field
predicates~\cite{langchainSelfQuery}. Attribute extraction~\cite{zou2024implicitave}, subjective
databases~\cite{li2019subjective}, review-based recommendation~\cite{mcauley2013hft}, IDSS~\cite{tran2026entropyguided}
and Cleo~\cite{schott2026cleo} rely on attribute inventories defined in advance. We define criteria from the current
request and assess them only for the retrieved candidates.

\section{Problem Formulation and Preliminaries}
\label{sec:background}

We study re-ranking after each user turn, when the active request may change and catalog records give incomplete
evidence; Table~\ref{tab:notation} lists the main notation.

\begin{table}[t]
\caption{Notation.}
\label{tab:notation}
\scriptsize
\setlength{\tabcolsep}{2.5pt}
\begin{tabular}{@{}ll@{\hspace{7pt}}ll@{}}
\toprule
Symbol & Meaning & Symbol & Meaning \\
\midrule
$i$ & candidate item & $m_t$, $P_t$ & message, pool at turn $t$ \\
$N$ & pool size & $\Phi_t$ & active features (needs) \\
$n$; $\phi$ & feature; inferred feature & $\Phi^{\mathrm{hard}}_t$, $\Phi^{\mathrm{soft}}_t$, $\Phi^{\mathrm{inf}}_t$ & hard, soft, inferred ones \\
$\sigma_n$ & polarity $\pm1$ & $\mathcal{D}$ & decision model \\
$p_s$, $p_c$, $p_u$ & \supp{}, \contra{}, \unk{} & $e$ & evidence $p_s-p_c$ \\
$T$, $S_F$ & hard tier; mean soft evidence & $E_H$, $S_{HF}$ & holistic, hybrid score \\
$g_i$ & grade of item $i$ & $W_t$ & pairs to decide at turn $t$ \\
\bottomrule
\end{tabular}
\end{table}

\subsection{Conversational re-ranking}
\label{sec:problem}
A catalog holds items $i$ with records $r_i$: fields (category, location, price) and free text (descriptions, reviews)
that vary across domains. A conversation consists of turns $m_1,m_2,\dots$. The active request after turn $t$ is a set
of needs $\Phi_t$, which may differ from earlier requests through additions, withdrawals, reversals or switches to
another request. A need $n$ has a predicate $\pi_n$ and a polarity $\sigma_n\in\{+1,-1\}$ for a wish or an avoidance.
The needs on the kind of item, its place and its price form the request's \emph{frame}, which a relevant item must
meet; the other needs are \emph{preferences} (``has a patio'', ``not a sports bar''). After each turn a first stage
retrieves a pool $P_t$ of $N$ candidates, and the recommender ranks $P_t$. Single-request recommendation is the case
$t=1$.

\paragraph{Target.} An item's grade $g_i$ states whether it truly meets $\Phi_t$: 0 if it fails the frame, 2 if it
meets the frame and every preference, and 1 if it meets the frame and part of the preferences (\S\ref{sec:benchmarks}
gives each dataset's rule). Truth comes from the item's attributes, which the recommender does not see: it must infer
need satisfaction from the remaining record, mostly its text. The task thus rewards ranking by true need
satisfaction. It does not test whether a per-need judgment is faithful to the record: a record can be silent on a need
that the item meets, and our grades cannot tell a correct ``insufficient evidence'' from a miss.

\subsection{Typed decision models}
\label{sec:dm}
A \emph{typed decision model} maps an input state $s$ and named questions $Q$ to option distributions,
\begin{equation}
\mathcal{D}\colon\;(s,Q)\;\mapsto\;\big(\mathbf{p}_q\big)_{q\in Q},
\label{eq:dm}
\end{equation}
where $\mathbf{p}_q$ is a distribution over the finite option set $O_q$ that question $q$ defines, with a criterion for
each option. Such models answer all
questions jointly in one forward pass without generating text, so one request can cover several records and their
questions. A generative LLM provides the same interface when $\mathbf{p}_q$ is read from the next-token probabilities of
the option names~\cite{zhuang2024finegrained}. \dr{} uses only the option probabilities, so either kind can serve as
$\mathcal{D}$. Every \dr{} question is a choice among named options: a
verdict needs ``insufficient evidence'' outside any order, and an ordered choice's expectation uses the same probabilities.

\section{The \drname{} Framework}
\label{sec:method}

\dr{} casts each turn as a set of typed decisions: one typed request reads the message and updates the feature state
(\S\ref{sec:features}, \S\ref{sec:state}), a typed decision model judges the candidates (\S\ref{sec:decide}), and one of
three scoring functions ranks them (\S\ref{sec:rank}; Figure~\ref{fig:overview}). No text is generated at query time.

\subsection{Typed interpretation}
\label{sec:features}
An interpreter $I$ maps the conversation, the catalog's fields $\mathcal{F}$ and the features defined so far to the
active feature set, $\Phi_t=I\big(m_{1:t},\,\mathcal{F},\,\Phi_{<t}\big)$. Each feature is one need. A \emph{schema
predicate} tests a field of $\mathcal{F}$, for example category contains ``Italian''; an
\emph{inferred feature} $\phi$ has a label and a definition that a record can support or contradict. Both
carry a polarity, so that ``touristy'' serves a wish and an avoidance alike. Strength follows a fixed rule: a
predicate on the item's type, location or price (the frame, \S\ref{sec:problem}), or one that excludes a field value,
is hard; all other needs are soft.

\dr{} implements $I$ with typed decisions instead of generation, in two steps. The \emph{vocabulary} $\mathcal{V}$
holds every (field, value) pair that the catalog's categorical fields contain, each yes/no field as one property.
\emph{Shortlist:} the message $m_t$ is split into clauses at punctuation and conjunctions; $m_t$ and each clause
retrieve their nearest entries of $\mathcal{V}$ by embedding similarity, and entries whose value occurs verbatim join
them, forming a shortlist $\mathcal{V}_t\subseteq\mathcal{V}$. \emph{Typed request:} one decision request asks, for
every entry $v\in\mathcal{V}_t$, whether $m_t$ asks for it, rules it out or neither,
\begin{equation}
\big(p_{\mathrm{want}},p_{\mathrm{avoid}},p_{\mathrm{neither}}\big)(v)=\mathcal{D}\big(m_t,\,Q_I(\mathcal{V}_t)\big)_{q_v},
\label{eq:stance}
\end{equation}
together with a bound on each ordered field and, for every clause, whether it asks for or rules out a property of the
item at all. A bound is one question whose options are limits drawn from the field's levels, plus ``no limit''; our
catalogs' only such field is the price level, so it asks for a price cap (\$, \$\$, \$\$\$ or none). The same question
would serve ratings or product dimensions, which no benchmark request constrains.
An entry wanted or avoided with probability at least $\frac12$ becomes a predicate, and the values of one field are
joined as alternatives. For a kind-of-item value, ``want'' means that the value names what the item \emph{is}, not a
property it should have: catalogs name kinds by properties (``Wi-Fi spot'', ``Health food restaurant''), and reading
``has wifi'' as such a kind would lose the need and widen the kind at once. Clause splitting and the removal of
greetings and request phrasing (``can you find me'') use fixed English rules (Appendix~\ref{app:typed}).

\paragraph{Needs without a field.} A clause that asks for something but maps to no entry has no field to test: the
catalog does not record the property, or, as in our benchmarks, its attribute is withheld. It becomes an inferred
feature defined by the user's own words: ``has a patio'' becomes the feature defined as ``The item has what these words
from the user describe: `has a patio'\,'', which the verifier judges from each record's text (\S\ref{sec:decide}).
Negation cues are removed and carried by the polarity. A clause that does map to entries is followed up by one more
question, whether it asks for anything about the item that those values do not state; if it does, its words become a
feature as well. A clause that names the kind of item is kept both ways. The kind's catalog values form the hard
predicate, which is coarse (``Seafood restaurant'' or ``Restaurant''), and the clause's words (``a seafood restaurant'')
become a soft feature that orders the candidates within it. Interpretation is thus schema-first, with the user's words
carrying what the schema states only coarsely or not at all. Such a feature's identity is its wording, so paraphrases
of one need are separate features.

\subsection{Need-level evidence}
\label{sec:decide}
A need's evidential relation to a record is
\begin{equation}
y(i,n)=\begin{cases}+1 & r_i \text{ supports } \pi_n,\\ -1 & r_i \text{ contradicts } \pi_n,\\
\phantom{+}0 & \text{evidence is absent or ambiguous}.\end{cases}
\label{eq:relation}
\end{equation}
For each candidate $i$ and inferred feature $\phi$, a verification question $q_\phi(i)$ has three options,
\supp{}, \contra{} and \unk{}, one per value of $y$:
\begin{equation}
\big(p_s,\,p_c,\,p_u\big)(i,\phi)\;=\;\mathcal{D}\big(s_V(B,\Phi^{\mathrm{inf}}_t),\,Q_V(B,\Phi^{\mathrm{inf}}_t)\big)_{q_\phi(i)},
\label{eq:verify}
\end{equation}
where the state $s_V$ holds a batch $B\ni i$ of up to $b$ records and the inferred feature definitions, and $Q_V$ holds
one question per candidate--feature pair. Schema predicates take one-hot distributions from direct field evaluation:
\supp{} when the predicate holds, \contra{} when it does not, and \unk{} when the value is absent. For either feature
type the evidence is the expected relation (Eq.~\ref{eq:relation}),
\begin{equation}
e(i,n)\;=\;\mathbb{E}\big[y(i,n)\big]\;=\;p_s(i,n)-p_c(i,n)\;\in[-1,1],
\label{eq:evidence}
\end{equation}
so \unk{} contributes nothing, while support and contradiction count with opposite signs. Ranking uses the full
distribution rather than the most likely option: even when \unk{} is most probable, $p_s-p_c$ can be nonzero.

\subsection{Three scoring functions}
\label{sec:rank}
\dr{} ranks a pool with one of three functions of typed decisions. They address the same task but differ in what they
ask per candidate, and hence in what they need and what they show.

\paragraph{Need-level (\drf{})} Hard needs form a tier, the sum of their signed evidence, and soft needs a mean,
\begin{equation}
T(i)=\sum_{n\in\Phi^{\mathrm{hard}}_t}\sigma_n\,e(i,n),\qquad
S_F(i)=\frac{1}{|\Phi^{\mathrm{soft}}_t|}\sum_{n\in\Phi^{\mathrm{soft}}_t}\sigma_n\,e(i,n),
\label{eq:key}
\end{equation}
and candidates are ordered by $T(i)$, then by $S_F(i)$, then by first-stage order. Each need counts once: a catalog
may state the kind of item in several fields (a product class and a product type), and the kind then contributes its
best evidence over them, so that a field one record leaves empty does not count against it. Every displayed candidate
carries a verdict per need, and a withdrawn or reversed need changes the ranking without new decisions; the mean,
however, treats needs as independent, and the ranking is only as good as the interpretation. The tier prioritizes
evidenced compliance with hard needs; it does not certify properties a record omits.

\paragraph{Holistic (\drh{})} One question $q_H(i)$ per candidate judges the whole request $x_t$---the message, the
dialogue or the feature state---with ordered options \emph{perfect} (all needs), \emph{mostly} (most needs, no
conflict), \emph{partly} (some needs or a conflict) and \emph{not}, valued $w=(3,2,1,0)$:
\begin{equation}
E_H(i)\;=\;\sum_{o}w_o\,\mathcal{D}\big(s_H(B,x_t),\,Q_H(B)\big)_{q_H(i),\,o},
\label{eq:holistic}
\end{equation}
where $s_H$ holds $x_t$ and a batch $B$ of records; candidates are ordered by $E_H(i)$. \drh{} needs no interpretation
of a single message and weighs needs jointly, but it gives no need-level evidence, has no hard tier, and must be asked
again for every candidate whenever the request changes.

\paragraph{Hybrid (\drhf{})} \drhf{} keeps \drf{}'s tier and orders the candidates within it by the holistic
decision and the need-level evidence together: by $T(i)$, then by
\begin{equation}
S_{HF}(i)\;=\;\tfrac13\,E_H(i)\;+\;S_F(i),
\label{eq:hybrid}
\end{equation}
where both terms lie on unit ranges ($E_H/3\in[0,1]$, $S_F\in[-1,1]$) and weigh equally. A hard need is never
outweighed. Within the tier, the holistic question weighs the needs jointly, which the verdicts cannot, and the
verdicts add per-need evidence, which the holistic question does not expose; neither weight is tuned.

\subsection{Turn transitions and decision routing}
\label{sec:state}
The typed request that interprets a turn also tracks the conversation. It asks whether the message changes the request
at all (a request, a question about the shown items, or an acknowledgement); whether it continues the current request,
starts a new one or returns to an earlier one; and, for every need defined so far, whether the message leaves it,
withdraws it or reverses it. The state update is deterministic: a non-request leaves the state as it is, a new request
starts empty, and a return restores the earlier request's needs with their identifiers; withdrawn needs leave the
state, reversed ones flip polarity under the same identifier, and the needs mapped from the message join.

Decisions are stored by candidate and feature identifier and remain valid while records and definitions are unchanged.
With $A_{<t}$ the pairs decided before turn $t$, a turn requires
\begin{equation}
W_t=\big\{(i,\phi)\in P_t\times\Phi^{\mathrm{inf}}_t:(i,\phi)\notin A_{<t}\big\}.
\label{eq:work}
\end{equation}
Each transition has a fixed effect: a withdrawn feature drops its term from Eq.~\ref{eq:key}, a reversed one flips
$\sigma_n$, and a restored one returns with its stored decisions. Only added features are judged on every candidate;
the others need decisions only for candidates new to the conversation, and a non-request turn needs none and keeps the
pool and the ranking. \drh{}'s question covers the whole request, read from the dialogue or from the typed state
rendered as a list of needs, so any request change asks it anew.

\section{Experimental Setup}
\label{sec:setup}
We evaluate single requests and synthetic conversations on two datasets with attribute-derived grades, with every
method re-ranking the same pools, to answer three research questions:
\begin{description}
\item[RQ1] How does ranking with typed decisions compare with schema filtering and LLM re-rankers?
\item[RQ2] Do typed decisions track request switches, revisions and returns?
\item[RQ3] What ranking-quality, latency and cost trade-offs result?
\end{description}

\begin{table}[t]
\caption{Dataset statistics. Words and full matches in the pool of 50 are per-request medians; conversations
with their turns.}
\label{tab:benchdata}
\footnotesize
\setlength{\tabcolsep}{4pt}
\begin{tabular}{@{}lrr@{}}
\toprule
 & GL & Yelp \\
\midrule
Items & 3,470 & 6,211 \\
Requests (test) & 320 & 320 \\
Needs: wished + avoided & 1--3 + 0--1 & 1--4 + 0--2 \\
Words per request & 20 & 27 \\
Full matches in pool & 3 & 4 \\
Conversations (turns) & 160 (6--8) & 160 (6--10) \\
\bottomrule
\end{tabular}

\end{table}

\subsection{Datasets}
\label{sec:benchmarks}
Table~\ref{tab:benchdata} summarizes the datasets; Figure~\ref{fig:benchmarks} illustrates their construction.
Google Local (GL)~\cite{li2022uctopic,yan2023showcases} holds food and drink places in Rhode Island, and Yelp open food
and nightlife businesses with at least 20 reviews in Pennsylvania~\cite{yelpOpenDataset2026}. Records contain the
usual listing fields, the listing's attributes (withheld, \S\ref{sec:benchmarks}) and the eight most recent
substantive reviews. Appendix~\ref{app:wands} adds product search on WANDS~\cite{chen2022wands}.

\paragraph{Requests and grades.} Each request is a specification that at least two catalog items fully meet: a place
type, optionally a town or price cap, and wished and avoided attributes. Yelp draws them from a real anchor business,
GL samples them and keeps those with two to \(\max(25, |\text{base}|/3)\) full matches. Gemini~2.5~Pro, a family no evaluated ranker uses,
writes each specification as one message and as two to four turns. On GL and Yelp an item is a full match (grade 2) if it has
the type, location and price, every wish and no avoided attribute, and a partial match (grade 1) if it has some wishes
and no avoided attribute. GL attributes are
positive-only; Yelp records explicit negative values (Appendix~\ref{app:bench}). All grades are
closed-world: an attribute the source does not record counts as absent, so a wish for it is unmet and an avoidance of it
met. Only 58\% and 78\% of pool grades on GL and Yelp stay the same whatever the
unrecorded attributes are; Appendix~\ref{app:labelpolicy} rescores every comparison
with them resolved against and for each item, and on the items whose grade is certain.

\paragraph{Withheld attributes.} Records withhold the graded attributes, which only the labels keep. Ranking thus
tests whether a method infers true need satisfaction from the remaining record (\S\ref{sec:problem}), not whether a
verdict is faithful to it. Withholding them also removes them from the interpreter's vocabulary: on GL and Yelp it holds
only kinds of place and towns or cities (341 and 487 entries), so every other need
takes the fallback of \S\ref{sec:features}, and 91\% and 86\% of
requests keep at least one need in the user's words.

\subsection{Conversations}
\label{sec:conversations}
We pair random requests A and B into 160 conversations per dataset: the user states A, switches to
B, revises B by dropping, replacing or reversing one need, and returns to A with its own revision, written and graded
like the requests. Rankings are scored at four
checkpoints: after A (P1), after B (P2), after B's revision (P3) and after the return (P4); we report P1 and the mean of
P2--4.

\paragraph{Dialogue versus typed state.} \emph{Dialogue} includes all user messages through the checkpoint, including superseded requests. The interpreter
reads the same history incrementally; its predicted active predicates, needs and polarities form the \emph{typed
state}. \drf{} scores this state. For \drh{}, we hold the model,
question and records fixed and replace the dialogue by the state rendered as a list of needs. For \drhf{},
\emph{typed state + dialogue} means \drf{} on the state plus \drh{} on the dialogue; \emph{typed state} feeds the state
to both halves. Neither receives gold needs or transitions.

\subsection{Candidate pools and inputs}
\label{sec:pools}
As is standard for comparing re-rankers on one first
stage~\cite{sun2023rankgpt,qin2024prp,zhuang2024setwise,pradeep2023rankzephyr}, every method re-ranks the same
$N{=}50$ candidates per request or checkpoint: reciprocal-rank fusion ($k{=}60$)~\cite{cormack2009rrf} of
BM25~\cite{robertson2009bm25} and dense retrieval (text-embedding-3-large) over the records as methods see them, for
the request text. No generative model takes part in retrieval. For conversations, the script groups
the current request's turns for retrieval, identically for all methods; the experiment isolates re-ranking, not
end-to-end retrieval under a predicted state. The pools are retrieved, not sampled, so the bias of sampled
metrics~\cite{krichene2020sampled} does not arise, but they bound every method: nDCG@10's ideal ranking covers every
graded item of the dataset, and we report the pool's best possible score with the results~\cite{wang2024birco}.
Decision models
receive records as JSON, each field cut to 3,000 characters (Appendix~\ref{app:setup}).

\subsection{Method configurations}
\label{sec:config}
\label{sec:backend-setup}
\paragraph{\dr{}.} A hosted typed decision model, Microsoft-Decision-1~\cite{msDecision1}, answers the typed
requests (one per message, a second when a clause needs the follow-up question; Appendix~\ref{app:typed}). It also answers the verification questions, for batches of $b{=}12$ candidates with a request's batches in parallel,
and \drh{}'s four-level question in batches of the same size. No model weights are adapted, and \drhf{} has no
parameter. The question templates and the composition and scoring rules were developed on development requests that
are disjoint from the benchmarks and built like them (places in Vermont and Florida, and held-out product queries), and were fixed before
any benchmark run (Appendix~\ref{app:typed}). Appendix~\ref{app:llminterp} replaces the typed request by an LLM interpreter that generates the
features.

\paragraph{Baselines.} Baselines receive the request (the dialogue in conversations) and the same pool. Input budgets
differ where a model's context requires it: the cross-encoder reads 512 tokens per record, the one-choice decision
baseline shortens records to fit one request (1,500 characters on Yelp), and the agent sees short cards but can read
full records; full configurations are in Appendix~\ref{app:setup}.
\emph{Conventional:} BM25, dense retrieval, pool order, bge-reranker-v2-m3~\cite{bgeRerankerV2M3ModelCard} and
self-query search~\cite{langchainSelfQuery}.
\emph{Pointwise:} Qwen3-Reranker-4B/-8B~\cite{zhang2025qwen3embedding}, GPT-4o and Qwen3.8-27B, the latter two scored
by expected four-level relevance~\cite{zhuang2024finegrained}.
\emph{Listwise:} GPT-4o through GPT-6 Sol/Luna~\cite{openai2024gpt4o,openai2025gpt5,openaiGPT6Sol,openaiGPT6Luna};
GPT-4o also uses RankGPT's sliding window~\cite{sun2023rankgpt}.
\emph{Agent/decision-oriented:} a GPT-4o tool agent~\cite{huang2023interecagent,zhao2024toolrec} and one hosted-model
choice over the pool~\cite{lyu2026decisionrerank}.
\emph{Conversational controls:} \drh{} and GPT-6 Sol on the raw dialogue; a stateful GPT-4o query
rewrite~\cite{elgohary2019canard,mao2023llm4cs} that also receives our state-update rules and its earlier rewrites,
scored by \drh{}.

\subsection{Evaluation}
\label{sec:protocol}
We report nDCG@10 with gains $2^{g}-1$ and the ideal ranking over all graded items of the dataset (\S\ref{sec:pools}).
Every method ranks the same pool, so methods are compared by their per-request differences with a paired bootstrap~\cite{bergkirkpatrick2012empirical,smucker2007significance} over requests, or over
whole conversations, Holm-corrected~\cite{holm1979simple} within each dataset and rechecked by a paired randomization
test (Appendix~\ref{app:tests}). GL and Yelp requests that share a kind of place and town or city are dependent, so
they are resampled and sign-flipped together. We call a difference significant only after correction and do not read a
non-significant difference as equivalence. Latency composes measured service times: parallel calls take their
maximum and sequential calls add, excluding the shared retrieval; costs apply list prices to logged tokens
(Appendix~\ref{app:latency}).

\section{Results}
\label{sec:results}
Table~\ref{tab:tests} gives the interval behind each claim, and Table~\ref{tab:labelpolicy}
repeats the main comparisons under other treatments of unrecorded attributes.

\subsection{Single requests (RQ1)}
\label{sec:main}

\begin{table}[t]
\caption{Single requests (nDCG@10). Bold: best; $^{*}$/$^{\circ}$: significantly worse/better than \drhf{}
(paired bootstrap, Holm, $p<0.05$).}
\label{tab:main}
\footnotesize
\setlength{\tabcolsep}{4pt}
\begin{tabular}{lcc}
\toprule
 & GL & Yelp \\
\midrule
\multicolumn{3}{@{}l}{\emph{Conventional}} \\
\quad Self-query, pool & 0.326\rlap{$^{*}$} & 0.381 \\
\quad bge-reranker-v2-m3 & 0.283\rlap{$^{*}$} & 0.341\rlap{$^{*}$} \\
\multicolumn{3}{@{}l}{\emph{Pointwise LLM}} \\
\quad Qwen3-Reranker-4B & 0.299\rlap{$^{*}$} & 0.364\rlap{$^{*}$} \\
\quad Qwen3-Reranker-8B & 0.309\rlap{$^{*}$} & 0.364\rlap{$^{*}$} \\
\quad Qwen3.8-27B, four levels & 0.354 & 0.427 \\
\multicolumn{3}{@{}l}{\emph{Listwise LLM}} \\
\quad GPT-4o & 0.313\rlap{$^{*}$} & 0.382\rlap{$^{*}$} \\
\quad GPT-5, minimal reasoning & 0.368 & 0.423 \\
\quad GPT-5 & 0.362 & 0.409 \\
\quad GPT-6 Luna & 0.349\rlap{$^{*}$} & 0.408 \\
\quad GPT-6 Sol & 0.343\rlap{$^{*}$} & 0.412 \\
\multicolumn{3}{@{}l}{\emph{Agentic}} \\
\quad GPT-4o agent & 0.310\rlap{$^{*}$} & 0.375\rlap{$^{*}$} \\
\multicolumn{3}{@{}l}{\emph{Decision-oriented reranking}} \\
\quad Hosted decision model & 0.338\rlap{$^{*}$} & 0.393\rlap{$^{*}$} \\
\multicolumn{3}{@{}l}{\emph{\dr{}}} \\
\rowcolor{oursbg}\quad \drh{} & 0.354 & 0.404 \\
\rowcolor{oursbg}\quad \drf{} & 0.308 & 0.343 \\
\rowcolor{oursbg}\quad \drhf{} & \textbf{0.383} & \textbf{0.428} \\
\bottomrule
\end{tabular}

\end{table}

\emph{Question.} Can typed decisions rank single requests as well as LLM rankers when the graded attributes are
withheld?

\emph{Design.} Every method re-ranks the same 50 candidates per request; \dr{} uses the hosted decision model without
adaptation. The pool's best ranking reaches 0.706 on GL and 0.736 on Yelp. Using only
type, town and price, with random ties, reaches 0.318 and 0.370; this is not a bound
on what free text can reveal.

\emph{Results.} \drhf{} scores 0.383 on GL and 0.428 on Yelp (Table~\ref{tab:main}). On GL it ranks better than GPT-6 Sol by $+$0.040 (CI
[$+$0.020, $+$0.063]), significant after requests sharing a kind of place and town are resampled together and the
comparisons are Holm-corrected, and it is not significantly different from the strongest baseline, GPT-5 at minimal
reasoning ($+$0.015, CI [$-$0.006, $+$0.039]). On Yelp it is not significantly different from
GPT-6 Sol ($+$0.016, CI [$-$0.002, $+$0.035]) or from the strongest baseline, Qwen3.8-27B
($+$0.001, CI [$-$0.016, $+$0.018]). The tier and the verdicts add
$+$0.029 on GL and $+$0.024 on Yelp to the holistic decision, both significant. The comparisons with \drh{} and GPT-6 Sol keep their sign
when unrecorded attributes are resolved against or for every item, or only grades that do not depend on them are scored
(Table~\ref{tab:labelpolicy}).

\emph{Analysis.} The holistic question carries the ranking; the tier removes candidates of the wrong kind, place or
price that it rates highly, and the verdicts order what it ties. Need-level verdicts alone rank low (\drf{}:
0.308 on GL, 0.343 on Yelp). This is partly a target mismatch: \drf{} rewards record
evidence, whereas the grades reward withheld attributes, so a correct \unk{} earns nothing. It is also an
interpretation limit: needs kept in the user's words verify worse than generated definitions, and an LLM interpreter
raises \drf{} to 0.391 and 0.444 and \drhf{} to 0.405 and 0.448, at the
cost of generated text (Appendix~\ref{app:llminterp}). The tier is only as good as its predicates: on product
search, where the typed request can map a query onto the wrong product type, a wrong hard predicate demotes every
relevant item (Appendix~\ref{app:wands}).
\emph{Typed decisions, combined must-first, rank local-business requests at least as well as the strongest LLM rankers
within our statistical resolution.}

\subsection{Conversations (RQ2)}
\label{sec:convo}

\begin{table}[t]
\caption{Conversations (160 per dataset): nDCG@10 after request A (P1), over the switch to B, its revision and the
return to A (P2--4), and the change $\Delta$. + dialogue: \drhf{}'s holistic half reads the dialogue. Bold: best;
$^{*}$: differs significantly from \drhf{}, typed state + dialogue (paired bootstrap over conversations, Holm,
$p<0.05$).}
\label{tab:convo}
\footnotesize
\setlength{\tabcolsep}{2.1pt}
\begin{tabular}{lcccccc}
\toprule
 & \multicolumn{3}{c}{GL} & \multicolumn{3}{c}{Yelp} \\
\cmidrule(lr){2-4} \cmidrule(lr){5-7}
 & P1 & P2--4 & $\Delta$ & P1 & P2--4 & $\Delta$ \\
\midrule
Pool ceiling (oracle order) & 0.704 & 0.682 & $-$0.022 & 0.729 & 0.701 & $-$0.029 \\
\midrule
GPT-6 Sol, dialogue & 0.336\rlap{$^{*}$} & 0.320 & $-$0.016 & 0.430 & \textbf{0.387}\rlap{$^{*}$} & $-$0.043 \\
Stateful rewrite + \drh{} & 0.360\rlap{$^{*}$} & \textbf{0.328} & $-$0.032 & 0.421\rlap{$^{*}$} & 0.370 & $-$0.051 \\
\midrule
\rowcolor{oursbg}\drh{}, dialogue & 0.354\rlap{$^{*}$} & 0.244\rlap{$^{*}$} & $-$0.110 & 0.424\rlap{$^{*}$} & 0.301\rlap{$^{*}$} & $-$0.123 \\
\rowcolor{oursbg}\drh{}, typed state & 0.339\rlap{$^{*}$} & 0.300\rlap{$^{*}$} & $-$0.038 & 0.376\rlap{$^{*}$} & 0.348 & $-$0.028 \\
\rowcolor{oursbg}\drf{}, typed state & 0.311\rlap{$^{*}$} & 0.291\rlap{$^{*}$} & $-$0.020 & 0.366\rlap{$^{*}$} & 0.327\rlap{$^{*}$} & $-$0.039 \\
\rowcolor{oursbg}\drhf{}, typed state + dialogue & \textbf{0.398} & 0.327 & $-$0.071 & \textbf{0.452} & 0.365 & $-$0.087 \\
\rowcolor{oursbg}\drhf{}, typed state & 0.351\rlap{$^{*}$} & 0.310 & $-$0.041 & 0.389\rlap{$^{*}$} & 0.359 & $-$0.030 \\
\bottomrule
\end{tabular}

\end{table}

\begin{table}[t]
\caption{The typed request's turn decisions scored against the script (\%).}
\label{tab:recog}
\footnotesize
\setlength{\tabcolsep}{3pt}
\begin{tabular}{@{}llcc}
\toprule
Decision & Scripted case (\%) & GL & Yelp \\
\midrule
Relation & switch read as a new request & 83 & 86 \\
 & return read as a return & 97 & 98 \\
\addlinespace[2pt]
Need & dropped need withdrawn & 19 & 27 \\
 & replaced value updated & 43 & 37 \\
 & reversal keeps the need & 27 & 26 \\
\bottomrule
\end{tabular}

\end{table}

\emph{Question.} Does an explicit typed state keep the ranking right as the user switches, revises and returns?

\emph{Design.} Table~\ref{tab:convo} uses the dialogue/state configurations of \S\ref{sec:conversations}
(logged state: Table~\ref{tab:statetrace}). The \drh{} contrast isolates the input representation; comparing
\drhf{} with a dialogue-only baseline also changes the scoring function. We score each turn decision against the script
(Table~\ref{tab:recog}; definitions in Appendix~\ref{app:typed}).

\emph{Results.} Replacing the dialogue by the typed state raises \drh{} over P2--4 by
$+$0.056 on GL (CI [$+$0.036, $+$0.077]) and
$+$0.047 on Yelp, both significant. At P1 it changes by
$-$0.015 and $-$0.048, significant only on Yelp. \drhf{} with the
dialogue in its holistic half has the best P1 on both datasets, above GPT-6 Sol by $+$0.063 on GL
(CI [$+$0.039, $+$0.088]). Over P2--4 it does not differ significantly from GPT-6 Sol on GL
($+$0.007, CI [$-$0.012, $+$0.026]) and is worse on Yelp
($-$0.022, CI [$-$0.039, $-$0.005]). The typed request finds nearly every
return and every switch between different kinds of place (92\% and
94\%), but withdraws exactly the dropped need, and nothing else, in only
19\% and 27\% of drops. The pool's own ceiling also
falls after the request changes (Table~\ref{tab:convo}), so part of every method's P2--4 decline comes from retrieval.

\emph{Analysis.} The state carries needs across switches and returns, but drops nuance, which costs \drh{} at P1.
Missed withdrawals leave dropped needs scoring: ``doesn't have to be good for kids'' is read as an avoidance, ``cash only
is fine'' as a wish. A switch between two requests for the same kind of place, 18 of
160 on GL and 15 on Yelp, gives the state no cue that a new request begins.
\emph{State maintenance helps after request changes, but interpretation loses nuance and misses most withdrawals.}

\subsection{Latency and cost (RQ3)}
\label{sec:latency}

\begin{table}[t]
\caption{Single requests: nDCG@10, median latency per request (s) and list-price cost per 1,000 requests (\$).
Bold: best. $^{\dagger}$: timed while the endpoint queued requests (Appendix~\ref{app:latency}).
$^{\ddagger}$: one A100, no list price.}
\label{tab:latency}
\footnotesize
\setlength{\tabcolsep}{3pt}
\begin{tabular}{lrrrrrr}
\toprule
 & \multicolumn{3}{c}{GL} & \multicolumn{3}{c}{Yelp} \\
\cmidrule(lr){2-4} \cmidrule(lr){5-7}
 & nDCG & s & \$/1k & nDCG & s & \$/1k \\
\midrule
\quad Qwen3.8-27B, four levels$^{\ddagger}$ & 0.354 & 7.0 & -- & 0.427 & 10.2 & -- \\
\quad Self-query, pool & 0.326 & 2.5 & 5.81 & 0.381 & --$^{\dagger}$ & 6.49 \\
\quad GPT-4o listwise & 0.313 & 3.5 & 49 & 0.382 & --$^{\dagger}$ & 81 \\
\quad GPT-5 listwise, minimal & 0.368 & 3.1 & 25 & 0.423 & 4.0 & 41 \\
\quad GPT-5 listwise & 0.362 & 80.2 & 78 & 0.409 & 57.5 & 93 \\
\quad GPT-6 Luna listwise & 0.349 & 10.8 & 2.56 & 0.408 & 18.6 & 3.94 \\
\quad GPT-6 Sol listwise & 0.343 & 5.6 & 43 & 0.412 & 6.3 & 69 \\
\quad GPT-4o agent & 0.310 & 15.2 & 71 & 0.375 & 16.7 & 103 \\
\rowcolor{oursbg}\quad \drh{} & 0.354 & \textbf{0.9} & \textbf{1.36} & 0.404 & \textbf{1.0} & \textbf{1.85} \\
\rowcolor{oursbg}\quad \drf{} & 0.308 & 2.5 & 1.64 & 0.343 & 2.8 & 2.09 \\
\rowcolor{oursbg}\quad \drhf{} & \textbf{0.383} & 2.5 & 3.00 & \textbf{0.428} & 2.8 & 3.94 \\
\bottomrule
\end{tabular}

\end{table}

\emph{Question and design.} What do latency and cost buy in ranking quality? Table~\ref{tab:latency} pairs the
results of RQ1 with the median latency per request and the cost of its logged tokens at public list prices. Latency
composes measured service times, excluding client-side waiting and the shared retrieval (\S\ref{sec:protocol}); it
is not throughput under load. We report medians, since tails depend on the endpoint's load when each run was made.

\emph{Results.} \drhf{} answers a request in 2.5\,s on GL and 2.8\,s on Yelp at
the median: interpretation (median 1.6\,s on GL, embedding included) and verification
(median 1.0\,s) run in sequence, and the holistic decision in parallel with them. GPT-6 Sol takes 5.6\,s and
6.3\,s, GPT-5 at minimal reasoning 3.1\,s and
4.0\,s, and GPT-5 at its default effort 80.2\,s and
57.5\,s. Per 1,000 requests, \drhf{} costs \$3.00 on GL and
\$3.94 on Yelp, against \$43 and
\$69 for GPT-6 Sol and \$25 and
\$41 for GPT-5 at minimal reasoning. GPT-6 Luna is the only LLM ranker in the
same price range (\$2.56 and \$3.94), at
10.8\,s and 18.6\,s. \drh{} alone is the fastest and
cheapest method in Table~\ref{tab:latency}, at 0.9\,s and 1.0\,s.

\emph{Analysis.} A decision request reads its batch of candidates and generates no text, so the decision
requests of \drh{} take about a second although together they read more input than a listwise prompt; the listwise
rankers' latency grows with the reasoning and ranking text they generate (Appendix~\ref{app:latency}). Typed
interpretation and verification add two sequential stages to \drh{}, which more than double its latency and cost and
buy the significant gain of RQ1. The cost gap reflects the decision model's tariff, which bills input only, as well as the
absence of generated output, so it is not an isolated effect of generation-free inference.
\emph{On these workloads, \drhf{} ranks at least as well as the strongest LLM rankers, with a lower median latency
than any of them and a fraction of their list-price cost, GPT-6 Luna's excepted.}

\section{Discussion and Limitations}
\label{sec:discussion}

\paragraph{Three functions, one set of decisions.} The holistic question carries the ranking. The must tier and the
need-level verdicts add to it when the frame is read correctly, and a wrong hard predicate costs more than they add,
as on product types (Appendix~\ref{app:wands}). The verdicts also give each displayed candidate need-level evidence and let
withdrawals and reversals change the ranking without new decisions (\S\ref{sec:convo}). Interpretation is the
bottleneck of both.

\subsection{Limitations}
\label{sec:limitations}
\paragraph{Benchmark validity.} ReDial~\cite{li2018redial}, INSPIRED~\cite{hayati2020inspired} and
Reddit-Movie~\cite{he2023llmcrs} lack need-level labels, so we generate requests and conversations from real records.
On human MultiWOZ dialogues~\cite{zang2020multiwoz}, the typed request reads 94\% of added
constraints and 88\% of replaced ones, but keeps fewer constraints than an LLM interpreter
(joint goal accuracy 0.701 against 0.817; Appendix~\ref{app:mwoz}). Grades lack human
audit and treat unrecorded attributes as absent, which on positive-only GL decides many negatives
(Appendix~\ref{app:labelpolicy}); the pools bound every method, and both datasets cover regional food and nightlife; product search is evaluated only
in Appendix~\ref{app:wands}.
The grades concern withheld attributes, so nDCG cannot validate record-level verdicts. Some scripted switches pair two
requests for the same kind of place, and shared conversational pools use scripted request boundaries, not predicted
ones. The turn decision on non-request messages is part of the method but not evaluated here. No user study tests the
evidence as an explanation.

\paragraph{Model and measurement scope.} Interpretation handles categorical fields, needs embeddings, and splits
clauses and removes filler with fixed English rules. Needs without a field keep the user's wording, so paraphrases do
not share a feature and leftover words can become needs (Appendix~\ref{app:typed}). Text-dependent needs are never hard, weights are equal, and long-term history is absent. We use one hosted
decision model without adaptation; training it from relevance is future work. Batching and re-querying change answers,
and latency is composed rather than measured under load.

\section{Conclusion}
\dr{} builds a conversational recommender from typed decisions. Ranked must-first, it ranks local-business requests at
least as well as strong LLM rankers within our statistical resolution, with lower median latency and at a fraction of
their cost, and its typed state keeps rankings across request switches and returns. Interpretation limits it: missed
withdrawals and wrong hard predicates are its main failures.

\section*{Ethical Considerations}
GL is research-licensed with citation and WANDS MIT-licensed; Yelp's terms prohibit sharing its data, so
we release build scripts. Adversarial descriptions can elicit \supp{} verdicts, avoided features can encode
discriminatory preferences, and unequal documentation can reveal properties; deployments should restrict harmful
criteria and treat record-level evidence as uncertain.

\section*{Acknowledgments}
An LLM assistant supported coding, experiments and drafting under the authors' direction; the authors reviewed all
code, results and text, and scripts generate every reported number from logged model outputs.

\bibliographystyle{ACM-Reference-Format}
\bibliography{refs_verified}

\appendix
\section{Typed Interpretation}
\label{app:typed}
\paragraph{Vocabulary.} For each dataset, $\mathcal{V}$ contains every value that occurs in at least two records of a
categorical field of the dataset's profile, excluding names, descriptions, reviews and hours: 341,
487 and 2360 entries on GL, Yelp and WANDS, whose records
withhold the graded attributes. A field whose values are only yes and no
becomes one property (``outdoor seating''); ``no'' and ``none'' values are dropped; price levels are asked as a cap. Each
entry is embedded once as ``field: value'' with text-embedding-3-large, the model used for dense retrieval.

\paragraph{Shortlist and questions.} A message is split at sentence ends, commas, dashes, conjunctions and relative
pronouns; a comma-separated piece with fewer than two content words joins the next piece (``a casual, laid-back
spot'' stays one clause). The message retrieves its 12 nearest entries and every clause its 8 nearest, with at most three (message) or
two (clause) per field, eight for the kind of item and two more for its place. Entries whose value occurs verbatim are
added, and at most 44 are kept. One typed request then asks three kinds of question:
\begin{itemize}
\item \emph{Entry grounding:} for every entry, whether the message asks for it (``exactly this property''; for kinds of
item, ``this value names the kind of item the message asks for (what the item is), or a narrower kind within it; not a
property, feature or service the message wants the item to have''), rules it out, or neither.
\item \emph{Price cap:} \$, \$\$, \$\$\$ or none, each described as a price level (``at most \$: the lowest price
level (cheap, inexpensive, low budget)''); asked only when the catalog's price field holds \$-levels.
\item \emph{Clause intent:} for every clause stripped of greetings and phrases such as ``can you find me'', whether it
asks for something about the item, rules something out, or says nothing about it.
\end{itemize}
Requests average 30, 40 and 23 questions on GL, Yelp
and WANDS, in at most 48 per decision request.

\paragraph{Composition.} An entry becomes a predicate when its most probable answer is ``want'' or ``avoid'' with
probability at least $\frac12$, keeping per clause at most four values of a field (eight for kinds of item) and two
avoided entries; the values of a field are joined as alternatives. A clause that asks for something and maps to no
entry becomes an inferred feature whose definition quotes its words (``The item has what these words from the user
describe: \dots{} (ignore words that only say how much the user wants it)''), with negation cues removed when the
clause rules something out. A clause from which a kind of item is mapped also becomes such a feature. When any other
clause maps to entries but has words they do not account for (``a \emph{classy} seafood restaurant''), a second typed
request asks whether the clause asks for anything about the item that those values do not state (``a quality, a
feature, a service, an amenity or a use'') or only names the kind of item, its place or those values; in the first
case the clause's words become a feature. Clauses that withdraw a need (``scratch the patio'') add none, and at most
four inferred features are kept, as for the LLM interpreter.

\paragraph{Needs kept in the user's words.} With the graded attributes withheld, most needs other than the kind of
item and the place have no entry and take the fallback of \S\ref{sec:features}. A feature is then identified by its
wording alone. On GL, the 24 requests that ask for outdoor seating give it
9 differently worded features (``has a patio'', ``we could sit outside'', ``outdoor
seating'', \dots) and none in 8; on Yelp, 19 wordings across
67 requests, and none in 31. Of all features, 14\%
on GL and 21\% on Yelp are single words, and 46 and
43 hold a region name (``rhode island'', ``casual Philly'') that the vocabulary, which lists
towns and cities, does not cover. A need can also be absorbed by a loosely similar kind of item: on Yelp, ``wheelchair
accessible'' is often read as asking for the category Venues \& Event Spaces, which joins the requested kinds as an
alternative, so the need is lost. Mapping needs onto the catalog's attribute schema, even where values are missing,
would give them stable identities; we leave it to future work.

\paragraph{Conversations.} At every turn after the first, the typed request also asks whether the message is a request,
a question about the shown items or an acknowledgement; whether it continues the current request, starts a new one or
returns to an earlier one (each listed by its first message); and, for every need defined so far, whether it is left,
withdrawn or reversed. Within a request, a newly mapped kind of item or place does not replace the current one; a new
kind is a new request. Like the LLM interpreter, the request reads the user's messages but not the items shown.

\paragraph{Turn decisions.} Recognition (Table~\ref{tab:recog}) scores each decision against the script: the
relation at the switch to B and the return to A, and the move the revision makes. A drop is correct only if exactly one
need, not the wished kind of item, place or price, is withdrawn and nothing else changes; a replacement if exactly one
such need is withdrawn and another added; a reversal if exactly one need flips and none is withdrawn. An avoided
category, such as ``no sports bars'' for ``no TVs'', counts as a need. Whether the moved need is the one the script
names is not checked.

\paragraph{A logged state.} Table~\ref{tab:statetrace} shows the typed state of one GL conversation after every turn,
one whose turn decisions all match the script: predicates on the kind of item and the town, needs kept in the user's
words, a new request that starts empty, a withdrawal, and a return that restores request A before withdrawing one of
its needs. \drf{} verifies the quoted needs and evaluates the predicates; \drh{} on the typed state reads the same
state rendered as a list of required and wished properties. Errors arise at the same decisions that
Table~\ref{tab:recog} scores. In another conversation, ``scratch the cozy vibe, how about a place with more of a college
crowd'' is read as a reversal of ``cozy'' rather than its replacement; ``really good coffee'' widens the kind of item to
coffee shops instead of becoming a need; and the filler ``About my first request: Actually'' becomes an avoided need.

\begin{table*}[t]
\caption{A logged conversation (GL): the typed request's decisions (every turn is read as a request) and
the typed state after every turn. Sans-serif: predicates on the catalog's fields; quoted: needs in the user's words, with negation cues removed and
carried by the polarity (avoid). Checkpoints P1--P4 are scored.}
\label{tab:statetrace}
\footnotesize
\begin{tabular}{@{}r@{\hspace{4pt}}l@{\hspace{6pt}}p{0.33\textwidth}@{\hspace{6pt}}p{0.17\textwidth}@{\hspace{6pt}}p{0.38\textwidth}@{}}
\toprule
 & & Message & Decision & State after the turn \\
\midrule
1 &  & A: hi i'm looking for a breakfast or brunch place in RI & new request (A) & \textsf{category: Breakfast restaurant, Brunch restaurant} \\
2 &  & A: somewhere with a really good tea selection & continues & \textsf{category: Breakfast restaurant, Brunch restaurant}; ``with a really good tea selection'' \\
3 & P1 & A: and happy hour deals would be a plus! & continues & \textsf{category: Breakfast restaurant, Brunch restaurant}; ``with a really good tea selection''; ``happy hour deals'' \\
4 &  & B: hey, looking for a coffee shop or cafe in newport. & new request (B) & \textsf{category: Coffee shop, Cafe, \dots; town: Newport} \\
5 &  & B: it's gotta have wifi. & continues & \textsf{category: Coffee shop, Cafe, \dots; town: Newport}; ``gotta have wifi'' \\
6 & P2 & B: and trying to avoid the college crowd, thanks. & continues & \textsf{category: Coffee shop, Cafe, \dots; town: Newport}; ``gotta have wifi''; avoid ``trying to the college crowd'' \\
7 & P3 & B: actually nvm about the student thing & continues; withdraws ``trying to the college crowd'' & \textsf{category: Coffee shop, Cafe, \dots; town: Newport}; ``gotta have wifi'' \\
8 & P4 & A: About my first request: actually never mind the happy hour part & returns to A; withdraws ``happy hour deals'' & \textsf{category: Breakfast restaurant, Brunch restaurant}; ``with a really good tea selection'' \\
\bottomrule
\end{tabular}

\end{table*}

\paragraph{Development.} The question templates, the composition rules and the scoring rules of
\S\ref{sec:rank} were developed on development requests disjoint from the benchmarks and built like them: GL requests
in Vermont, Yelp requests in Florida and WANDS requests over other original queries, 100 drawn at random
from each. They were fixed before any benchmark run. Development showed the failures the templates address: kind
values that name a property (``Wi-Fi spot'') absorbing the need; a follow-up question that let a mapped kind ``cover''
``outdoor seating''; leftover words joined into features (``Vermont selection''); and, on WANDS, records missing one
of two kind fields. Table~\ref{tab:dev} reports the configuration on these requests and the ablations of its design
choices. Interpretation needs one decision request, a second one for 63\%, 90\%
and 79\% of single requests, and one embedding call.

\begin{table}[h]
\caption{Development requests, disjoint from the benchmarks (nDCG@10). Indented rows change one design choice of
\drhf{}. $^{*}$/$^{\circ}$: worse/better than \drhf{} (paired bootstrap, $p<0.05$).}
\label{tab:dev}
\footnotesize
\begin{tabular}{lrrr}
\toprule
Configuration & GL & Yelp & WANDS \\
\midrule
\drh{} & 0.364 & 0.401 & 0.402 \\
\drf{} & 0.388 & 0.440 & 0.385 \\
\rowcolor{oursbg}\drhf{} & 0.421 & 0.447 & 0.404 \\
\quad within the tier, $E_H$ first, $S_F$ breaking its ties & 0.410 & 0.440 & 0.385 \\
\quad the kind counted once per field & 0.421 & 0.447 & 0.355\rlap{$^{*}$} \\
\quad the kind not kept in the user's words & -- & -- & -- \\
\quad verification: ``supports'' / ``explicitly conflicts'' & 0.412 & -- & -- \\
\bottomrule
\end{tabular}

\end{table}

\section{LLM Interpreter}
\label{app:llminterp}
\paragraph{Configuration.} The LLM interpreter replaces the typed request (\S\ref{sec:features}, \S\ref{sec:state}) by
a generative LLM, GPT-4o at temperature 0, which receives the dialogue, the catalog's field names and frequent values,
and its earlier features, and returns schema predicates and at most four inferred features, each with an operational
definition (Appendix~\ref{app:prompts}). It also produces the predicates of the self-query baseline and the features
of the option-name study (Appendix~\ref{sec:verdicts}).

\paragraph{Single requests.} The LLM interpreter raises \drf{} to 0.391,
0.444 and 0.477, and \drhf{} by $+$0.022 and $+$0.021 on GL
and Yelp (Table~\ref{tab:latent}). Table~\ref{tab:interplatent} isolates interpretation in \drf{}: typed
interpretation trails the LLM on all three datasets ($-$0.083, $-$0.101 and
$-$0.046), although its hard predicates keep 94\%, 88\%
and 66\% of the relevant candidates. The loss is in the needs no field expresses, which here
include every graded attribute: kept in the user's words, they rank far below the LLM's operational definitions on GL
and Yelp.

\begin{table}[h]
\caption{Interpretation for \drf{} (nDCG@10, single requests). Fields only and words only: the typed
states with the other part removed. $^{*}$/$^{\circ}$: typed better/worse than the LLM interpreter (paired bootstrap,
Holm, $p<0.05$); bold: best.}
\label{tab:interplatent}
\footnotesize
\begin{tabular}{lrrr}
\toprule
Interpretation & GL & Yelp & WANDS \\
\midrule
LLM interpreter & \textbf{0.391} & \textbf{0.444} & \textbf{0.477} \\
\rowcolor{oursbg}Typed interpretation & 0.308\rlap{$^{\circ}$} & 0.343\rlap{$^{\circ}$} & 0.431\rlap{$^{\circ}$} \\
\rowcolor{oursbg}\quad fields only & 0.296 & 0.366 & 0.410 \\
\rowcolor{oursbg}\quad user's words only & 0.149 & 0.219 & 0.462 \\
\bottomrule
\end{tabular}

\end{table}

\paragraph{What the LLM interpreter adds.} Where needs are values the catalog holds, typed decisions read them as well
as the LLM; the LLM adds an operational definition for a need the catalog does not express, which a decision model
verifies more reliably than the user's own words, and on WANDS also the product types. Typed interpretation thus
fits catalogs with rich fields and latency-bound deployments, and an LLM interpreter open-ended requests.

\section{Human Dialogues (MultiWOZ)}
\label{app:mwoz}
Our requests and conversations are generated. To check the typed request on messages people wrote, we run it,
untuned, on MultiWOZ~2.2~\cite{zang2020multiwoz}, human--human dialogues at a Cambridge information desk. The
117 restaurant-only dialogues of its test and validation splits (546 user turns) are
read turn by turn, with the desk's replies in the conversation so far, by the typed request with the per-turn questions
and deterministic update of \S\ref{sec:state}, and by the LLM interpreter of Appendix~\ref{app:llminterp}. The
catalog is MultiWOZ's restaurant database (110 records with name, food, area and price range). We score the state on
food, area and price range against the annotated dialogue state, with ``don't care'' as no constraint. The annotation
also takes in values the desk offered and the user merely accepted, so a value counts only once the user has written
it or a synonym; the 62 slot-turns whose value the user never wrote are not scored. A slot is read when
a wanted predicate on its field has the annotated value or, without one, a need kept in the user's words names it.

The typed request recognizes 94\% of the 259 slots the user adds and
88\% of the 34 values the user replaces, against
97\% and 91\% for the LLM interpreter, and drops 5
of the 6 slots the user drops (LLM: 4). Its joint accuracy over the scored
slots is lower, 0.701 against 0.817 (difference $-$0.115, 95\% interval [$-$0.182, $-$0.050],
bootstrap over dialogues), because a constraint it loses stays lost: in 27 turns it withdraws a
food, area or price that the user keeps, typically while another attribute changes (``How about an expensive British
place?'') or when ``any restaurant available'' with a booking reads as a fresh start (6
turns read as a new request). It also keeps booking details in the user's words. These dialogues contain few
withdrawals, so they test mapping and replacement on human wording rather than the withdrawals on which typed
decisions fall short in our conversations.

\section{Decision Model}
\label{app:backends}
\label{app:landscape}
\paragraph{Typed decision models} TypeSafe's Jev defined the interface that the models we use share: a request carries a
state and named questions, each with instructions and options described by criteria, and the response returns a
probability for every option and no generated text~\cite{typesafeSystemOne2026,typesafeJev2026}. Microsoft-Decision-1
serves the same request~\cite{msDecision1}, and OpenAI's Decisions API a similar one~\cite{openaiDecisions2026}; open
alternatives include Clef~\cite{cloudflareClef} and training-free readouts of generative
models~\cite{mapikaDecider2026,nokiaAnyJev2026}. We use Microsoft-Decision-1 (hosted, post-trained from
Qwen3.5-9B~\cite{msDecision1Announcement2026}) and Clef (open weights, 27B, post-trained from Qwen3.8-27B with a joint
decision head); Jev and the Decisions API were not available to our account. Most of what is known about these models
comes from vendor documentation and recent preprints that have not been peer reviewed.

\section{Statistical Tests}
\label{app:tests}
Every method ranks the same pool for a request, so two methods are compared by their per-request differences in
nDCG@10. The paired bootstrap resamples the requests of a dataset with replacement 10,000 times; the 2.5th and 97.5th
percentiles of the resampled mean difference give the 95\% interval, and the two-sided $p$-value is twice the smaller
share of resampled means on either side of zero, i.e.\ the smallest level at which the interval excludes
zero~\cite{bergkirkpatrick2012empirical,smucker2007significance}. In conversations the unit is the conversation, whose
checkpoints are dependent: the statistic is its P1 score or its mean over P2--4. Within each dataset, the comparisons of one configuration against all baselines
form a family whose $p$-values are Holm-corrected~\cite{holm1979simple}. Bootstrap tests can be anti-conservative on
small samples~\cite{urbano2019significance}, so a paired randomization test, which flips the sign of each request's
difference at random (on GL and Yelp, the summed difference of each cluster of dependent requests), rechecks every
corrected decision of Table~\ref{tab:main}; 137 of
141 agree. The bootstrap resamples whole clusters of GL and Yelp requests that share a kind of place and town or city (80 and 64
clusters), since such requests share candidates. We report no equivalence claims: a margin chosen from the same data
only restates the interval. The intervals cover the sampling of requests and conversations only: the datasets, the request writer and model nondeterminism are held fixed. Table~\ref{tab:tests} lists the differences behind the claims of \S\ref{sec:results}.

\begin{table*}[h]
\caption{Differences behind the claims of \S\ref{sec:results}: mean difference and 95\% paired-bootstrap interval.}
\label{tab:tests}
\footnotesize
\setlength{\tabcolsep}{4pt}
\begin{tabular}{@{}llll@{}}
\toprule
 & GL & Yelp & WANDS \\
\midrule
\multicolumn{4}{@{}l}{\emph{Single requests (nDCG@10)}} \\
\drhf{} $-$ GPT-6 Sol & $+$0.040\,[$+$0.020, $+$0.063] & $+$0.016\,[$-$0.002, $+$0.035] & $-$0.099\,[$-$0.129, $-$0.070] \\
\drhf{} $-$ strongest baseline$^{a}$ & $+$0.015\,[$-$0.006, $+$0.039] & $+$0.001\,[$-$0.016, $+$0.018] & $-$0.099\,[$-$0.129, $-$0.070] \\
\drhf{} $-$ \drh{} & $+$0.029\,[$+$0.018, $+$0.041] & $+$0.024\,[$+$0.017, $+$0.033] & $-$0.070\,[$-$0.094, $-$0.047] \\
\multicolumn{4}{@{}l}{\emph{Conversations (nDCG@10 at P1 / mean over P2--4)}} \\
\drh{}: typed state $-$ dialogue, P1 & $-$0.015\,[$-$0.038, $+$0.009] & $-$0.048\,[$-$0.074, $-$0.024] & -- \\
\drh{}: typed state $-$ dialogue, P2--4 & $+$0.056\,[$+$0.036, $+$0.077] & $+$0.047\,[$+$0.028, $+$0.066] & -- \\
\multicolumn{4}{@{}l}{\emph{\quad\drhf{}, typed state + dialogue, $-$ baseline}} \\
GPT-6 Sol, dialogue, P1 & $+$0.063\,[$+$0.039, $+$0.088] & $+$0.021\,[$+$0.000, $+$0.043] & -- \\
GPT-6 Sol, dialogue, P2--4 & $+$0.007\,[$-$0.012, $+$0.026] & $-$0.022\,[$-$0.039, $-$0.005] & -- \\
Stateful rewrite + \drh{}, P2--4 & $-$0.001\,[$-$0.018, $+$0.016] & $-$0.005\,[$-$0.022, $+$0.014] & -- \\
\drh{}, dialogue, P2--4 & $+$0.083\,[$+$0.066, $+$0.100] & $+$0.064\,[$+$0.049, $+$0.080] & -- \\
\bottomrule
\multicolumn{4}{@{}l}{$^{a}$GPT-5 at minimal reasoning on GL, Qwen3.8-27B (four levels) on Yelp, GPT-6 Sol on WANDS.}
\end{tabular}
\end{table*}

\section{Additional Ranking Results}
\label{app:variants}
Table~\ref{tab:latent} extends Table~\ref{tab:main} with WANDS (Appendix~\ref{app:wands}), P@10, BM25, dense
retrieval, the pool order, catalog-wide self-query search, and \drf{} and \drhf{} with the LLM interpreter
(Appendix~\ref{app:llminterp}).

\begin{table*}[h]
\caption{All single-request variants (nDCG@10 and P@10). $^{*}$/$^{\circ}$: worse/better than \drhf{} (Holm, $p<0.05$); bold: best; --: not run.}
\label{tab:latent}
\small
\begin{tabular}{lcccccc}
\toprule
 & \multicolumn{2}{c}{GL} & \multicolumn{2}{c}{Yelp} & \multicolumn{2}{c}{WANDS} \\
\cmidrule(lr){2-3} \cmidrule(lr){4-5} \cmidrule(lr){6-7}
 & nDCG@10 & P@10 & nDCG@10 & P@10 & nDCG@10 & P@10 \\
\midrule
\multicolumn{7}{@{}l}{\emph{Conventional}} \\
\quad BM25 & 0.107\rlap{$^{*}$} & 0.057 & 0.173\rlap{$^{*}$} & 0.087 & 0.373\rlap{$^{*}$} & 0.194 \\
\quad Dense retrieval & 0.268\rlap{$^{*}$} & 0.143 & 0.320\rlap{$^{*}$} & 0.166 & 0.439\rlap{$^{*}$} & 0.221 \\
\quad Pool order (first stage) & 0.233\rlap{$^{*}$} & 0.118 & 0.307\rlap{$^{*}$} & 0.163 & 0.434\rlap{$^{*}$} & 0.209 \\
\quad Self-query, catalog-wide & 0.356 & 0.188 & 0.383 & 0.198 & -- & -- \\
\quad Self-query, pool & 0.326\rlap{$^{*}$} & 0.166 & 0.381 & 0.200 & 0.443 & 0.217 \\
\quad bge-reranker-v2-m3 & 0.283\rlap{$^{*}$} & 0.149 & 0.341\rlap{$^{*}$} & 0.183 & 0.496 & 0.271 \\
\multicolumn{7}{@{}l}{\emph{Pointwise LLM}} \\
\quad Qwen3-Reranker-4B & 0.299\rlap{$^{*}$} & 0.162 & 0.364\rlap{$^{*}$} & 0.194 & 0.548\rlap{$^{\circ}$} & 0.307 \\
\quad Qwen3-Reranker-8B & 0.309\rlap{$^{*}$} & 0.166 & 0.364\rlap{$^{*}$} & 0.195 & 0.548\rlap{$^{\circ}$} & 0.311 \\
\quad Qwen3.8-27B, four levels & 0.354 & 0.190 & 0.427 & \textbf{0.240} & 0.553\rlap{$^{\circ}$} & 0.324 \\
\multicolumn{7}{@{}l}{\emph{Listwise LLM}} \\
\quad GPT-4o & 0.313\rlap{$^{*}$} & 0.166 & 0.382\rlap{$^{*}$} & 0.199 & 0.542\rlap{$^{\circ}$} & 0.300 \\
\quad GPT-5, minimal reasoning & 0.368 & 0.190 & 0.423 & 0.227 & 0.552\rlap{$^{\circ}$} & 0.306 \\
\quad GPT-5 & 0.362 & 0.198 & 0.409 & 0.232 & 0.579\rlap{$^{\circ}$} & 0.338 \\
\quad GPT-6 Luna & 0.349\rlap{$^{*}$} & 0.190 & 0.408 & 0.228 & 0.581\rlap{$^{\circ}$} & 0.342 \\
\quad GPT-6 Sol & 0.343\rlap{$^{*}$} & 0.193 & 0.412 & 0.233 & \textbf{0.588}\rlap{$^{\circ}$} & \textbf{0.351} \\
\multicolumn{7}{@{}l}{\emph{Agentic}} \\
\quad GPT-4o agent & 0.310\rlap{$^{*}$} & 0.166 & 0.375\rlap{$^{*}$} & 0.202 & 0.514 & 0.271 \\
\multicolumn{7}{@{}l}{\emph{Decision-oriented reranking (one choice over all candidates)}} \\
\quad Hosted decision model & 0.338\rlap{$^{*}$} & 0.176 & 0.393\rlap{$^{*}$} & 0.212 & 0.541\rlap{$^{\circ}$} & 0.307 \\
\multicolumn{7}{@{}l}{\emph{\dr{}}} \\
\rowcolor{oursbg}\quad \drh{} & 0.354 & 0.195 & 0.404 & 0.220 & 0.559 & 0.330 \\
\rowcolor{oursbg}\quad \drf{} & 0.308 & 0.172 & 0.343 & 0.192 & 0.431 & 0.230 \\
\rowcolor{oursbg}\quad \drhf{} & 0.383 & 0.209 & 0.428 & 0.230 & 0.489 & 0.276 \\
\rowcolor{oursbg}\quad \drhf{}, LLM interpreter & \textbf{0.405} & \textbf{0.214} & \textbf{0.448} & 0.238 & 0.548 & 0.325 \\
\rowcolor{oursbg}\quad \drf{}, LLM interpreter & 0.391 & 0.205 & 0.444 & 0.237 & 0.477 & 0.267 \\
\bottomrule
\end{tabular}

\end{table*}

\section{Product Search (WANDS)}
\label{app:wands}
\paragraph{Setup.} WANDS~\cite{chen2022wands} holds 42,994 furniture and home products whose records
contain name, class, description and attribute fields. Each of its 190 requests adds to an
original query 0--2 wished and 0--1 avoided yes/no attributes that split the
query's exact matches (Appendix~\ref{app:bench}); a request has a median of 12 words and
5 full matches in its pool. The grade is the WANDS grade minus one per unmet injected need.
WANDS records explicit negative values, and 58\% of pool grades stay the same whatever the
unrecorded attributes are (Appendix~\ref{app:labelpolicy}). Records withhold the 30 injectable attributes but keep
their other product fields, so the typed vocabulary holds 2360 entries. Pools fuse BM25 and BGE-large
retrieval for the original query, and the 14\% of pool items without a WANDS judgment count as
irrelevant. Requests come from distinct queries and are resampled independently; everything else follows the main
experiments.

\paragraph{Results.} Table~\ref{tab:latent} reports every method. The pool's best ranking reaches
0.854, and the original grade alone, with random ties, 0.546.
\drhf{} scores 0.489: it trails the strongest baseline, GPT-6 Sol, by
$-$0.099 (CI [$-$0.129, $-$0.070]), and the tier and the verdicts change
\drh{}'s 0.559 by $-$0.070 (CI [$-$0.094, $-$0.047]), both significant.
\drhf{} answers in 2.4\,s at the median, against 5.9\,s
for GPT-6 Sol, and costs \$2.89 per 1,000 requests, against
\$3.47 for GPT-6 Luna, which scores 0.581
(Table~\ref{tab:latencylatent}).

\paragraph{Analysis.} The typed request often maps a product query onto the wrong product type (``edge chair mat''
onto chairs), and a wrong hard predicate demotes every relevant item. With the LLM interpreter, \drhf{} reaches
0.548, close to \drh{}. \emph{On product search, errors in the tier cost more than it adds, and
LLM listwise rankers stay ahead.}

\section{Cost and Latency Details}
\label{app:latency}
Table~\ref{tab:latencylatent} reports composed latency, token usage and estimated cost per request. Estimates use
cached response usage and public list prices, not measured production billing or finite-capacity throughput. The
single-request runs of \drh{} shared the decision endpoint with many concurrent jobs, whose queueing entered the
recorded service times; their latencies come from re-sending every request's decision requests with caches off on an
otherwise idle endpoint, which leaves the answers and rankings unchanged. The interpretation and verification
requests of \drf{} keep their recorded service times. Its verification requests take 0.26\,s and
0.21\,s per 10,000 input tokens on GL and Yelp at the median, as fast as the idle-endpoint
holistic requests (0.29\,s and 0.23\,s), and its smaller interpretation
requests take a median 0.6\,s and 0.6\,s each, so neither includes
comparable queueing. Service times exclude client-side waiting and retried attempts. On Yelp, the GPT-4o calls of the LLM
interpreter (and so of self-query search) and of GPT-4o listwise include endpoint queueing (GPT-4o listwise takes
37.2\,s at the median for a ranking of about 40 tokens, against 3.5\,s
on GL); they are not comparable and Table~\ref{tab:latency} omits them. Qwen3-Reranker-8B ran batched, without
per-request timing.

\begin{table*}[h]
\caption{Latency, tokens and cost per request; models run on our A100 are not priced (--); bold: lowest.}
\label{tab:latencylatent}
\small
\setlength{\tabcolsep}{2.4pt}
\begin{tabular}{lrrrrrrrrr}
\toprule
 & \multicolumn{3}{c}{GL} & \multicolumn{3}{c}{Yelp} & \multicolumn{3}{c}{WANDS} \\
\cmidrule(lr){2-4} \cmidrule(lr){5-7} \cmidrule(lr){8-10}
 & s & k tok & \$/1k & s & k tok & \$/1k & s & k tok & \$/1k \\
\midrule
\multicolumn{10}{@{}l}{\emph{Open models on one A100 (no list price)}} \\
\quad bge-reranker-v2-m3 & 2.0 & -- & -- & \textbf{0.6} & -- & -- & \textbf{0.6} & -- & -- \\
\quad Qwen3-Reranker-4B & 2.2 & -- & -- & 10.1 & -- & -- & 11.9 & -- & -- \\
\quad Qwen3.8-27B, four levels & 7.0 & -- & -- & 10.2 & -- & -- & 10.3 & -- & -- \\
\multicolumn{10}{@{}l}{\emph{Hosted models}} \\
\quad Self-query, pool & 2.5 & 2 & 5.81 & 41.2 & 2 & 6.49 & 3.0 & 6 & 17 \\
\quad GPT-4o listwise & 3.5 & 20 & 49 & 37.2 & 32 & 81 & 12.4 & 29 & 74 \\
\quad GPT-5 listwise, minimal & 3.1 & 20 & 25 & 4.0 & 32 & 41 & 3.7 & 29 & 37 \\
\quad GPT-5 listwise & 80.2 & 25 & 78 & 57.5 & 37 & 93 & 90.1 & 34 & 83 \\
\quad GPT-6 Luna listwise & 10.8 & 21 & 2.56 & 18.6 & 34 & 3.94 & 14.5 & 30 & 3.47 \\
\quad GPT-6 Sol listwise & 5.6 & 20 & 43 & 6.3 & 33 & 69 & 5.9 & 30 & 63 \\
\quad GPT-4o agent & 15.2 & 28 & 71 & 16.7 & 40 & 103 & 14.9 & 36 & 93 \\
\multicolumn{10}{@{}l}{\emph{\dr{}}} \\
\rowcolor{oursbg}\quad \drh{} & \textbf{0.9} & 32 & \textbf{1.36} & 1.0 & 44 & \textbf{1.85} & 1.0 & 36 & 1.53 \\
\rowcolor{oursbg}\quad \drf{} & 2.5 & 39 & 1.64 & 2.8 & 50 & 2.09 & 2.4 & 32 & \textbf{1.36} \\
\rowcolor{oursbg}\quad \drhf{} & 2.5 & 71 & 3.00 & 2.8 & 94 & 3.94 & 2.4 & 69 & 2.89 \\
\rowcolor{oursbg}\quad \drhf{}, LLM interpreter & 3.5 & 74 & 8.84 & 43.6 & 101 & 11 & 3.8 & 76 & 20 \\
\bottomrule
\end{tabular}

\end{table*}

The prices are GPT-4o \$2.50/\$10, GPT-5 \$1.25/\$10 and GPT-4.1-mini \$0.40/\$1.60 per million input/output
tokens~\cite{openaiPricing2026}, GPT-6 Sol \$2/\$10 and GPT-6 Luna \$0.10/\$0.50~\cite{openaiGPT6Sol,openaiGPT6Luna}, with
reasoning tokens counted as output, and Microsoft-Decision-1 \$0.042 per million input tokens, with output
free~\cite{msDecision1Announcement2026}. Models run on our A100 are not priced. With typed interpretation, \drhf{}
costs \$3.00 per 1,000 GL requests, all of it decision-model input apart from one embedding call; with
the LLM interpreter, interpretation dominates and \drhf{} costs \$8.84.

\begin{figure*}[h]
\centering
\includegraphics[width=\textwidth]{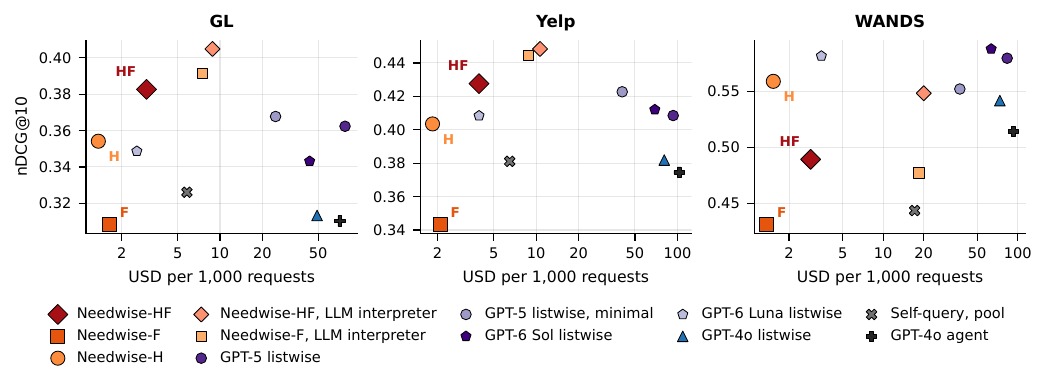}
\caption{Ranking quality versus list-price cost on single requests. Each point is a method's mean
nDCG@10 and token cost per 1,000 requests; open GPU models have no comparable list price and are omitted.}
\label{fig:costquality}
\end{figure*}

\paragraph{Where the time goes} A decision request reads its batch of candidates and generates no text, so \drh{}'s
decision requests take about a second, although together they read more input than any listwise prompt. GPT-5 at its
default effort writes a median of 5,120 reasoning tokens before its 55-token
ranking on GL, at 108 tokens per second where the endpoint reports its own timing; an independent
benchmark measures GPT-5 at medium effort on the same API at 111 tokens per second~\cite{aaGPT5Medium}.
Reasoning-token generation accounts for a substantial component of observed latency; the public throughput figure
is context, not an endpoint-matched measurement. The open re-rankers are timed per request alone on the A100
(Qwen3.8-27B on 30 requests). No latency includes the first stage, which every method shares.

\section{Feature-Level Verification}
\label{sec:verdicts}

\begin{table*}[t]
\caption{Per-need decisions on the per-need pairs; bold: best. ECE$_{\mathrm{com}}$ and Acc.$_{\mathrm{com}}$: over committed
answers; a label-only judge's ECE$_{\mathrm{com}}$ is its error rate.}
\label{tab:needsfull}
\label{tab:needs}
\footnotesize
\setlength{\tabcolsep}{4pt}
\begin{tabular}{@{}lccccccc@{}}
\toprule
Verifier & AUC & Prec. & Rec. & FSR & \unk{} & Acc.$_{\mathrm{com}}$ & ECE$_{\mathrm{com}}$ \\
\midrule
\multicolumn{8}{@{}l}{\emph{GL}} \\
\quad Keyword match & 0.583 & 0.757 & 0.244 & 8\% & 84\% & -- & -- \\
\quad Embedding threshold & 0.690 & 0.557 & 0.921 & 73\% & 17\% & -- & -- \\
\quad GPT-4o, label & 0.607 & 0.799 & 0.265 & 7\% & 80\% & 0.780 & 0.220 \\
\quad GPT-4o, token probabilities & 0.728 & 0.818 & 0.281 & 6\% & 76\% & 0.790 & 0.197 \\
\quad GPT-4o, stated probabilities & 0.604 & 0.713 & 0.259 & 10\% & 79\% & 0.723 & 0.217 \\
\quad Qwen3.8-27B, label & 0.601 & 0.837 & 0.229 & 4\% & 82\% & 0.804 & 0.196 \\
\quad Qwen3.8-27B, token probabilities & 0.690 & 0.818 & 0.296 & 7\% & 77\% & 0.785 & 0.150 \\
\quad Microsoft-Decision-1 & 0.726 & 0.860 & 0.153 & \textbf{2\%} & 91\% & \textbf{0.852} & 0.121 \\
\quad Clef (open, 27B) & \textbf{0.748} & 0.879 & 0.227 & 3\% & 82\% & 0.846 & \textbf{0.108} \\
\multicolumn{8}{@{}l}{\emph{Yelp}} \\
\quad Keyword match & 0.530 & 0.540 & 0.402 & 34\% & 63\% & -- & -- \\
\quad Embedding threshold & 0.646 & 0.507 & 0.994 & 97\% & 2\% & -- & -- \\
\quad GPT-4o, label & 0.670 & 0.825 & 0.370 & 8\% & 72\% & 0.832 & 0.168 \\
\quad GPT-4o, token probabilities & \textbf{0.777} & 0.813 & 0.411 & 9\% & 62\% & 0.810 & 0.181 \\
\quad GPT-4o, stated probabilities & 0.658 & 0.765 & 0.319 & 10\% & 73\% & 0.785 & 0.189 \\
\quad Qwen3.8-27B, token probabilities & 0.713 & 0.767 & 0.439 & 13\% & 65\% & 0.777 & 0.158 \\
\quad Microsoft-Decision-1 & 0.742 & 0.901 & 0.211 & \textbf{2\%} & 88\% & \textbf{0.900} & \textbf{0.077} \\
\quad Clef (open, 27B) & \textbf{0.777} & 0.857 & 0.310 & 5\% & 74\% & 0.854 & 0.093 \\
\multicolumn{8}{@{}l}{\emph{WANDS}} \\
\quad Keyword match & 0.666 & 0.705 & 0.572 & 24\% & 59\% & -- & -- \\
\quad Embedding threshold & 0.647 & 0.529 & 0.952 & 85\% & 10\% & -- & -- \\
\quad GPT-4o, label & 0.764 & 0.887 & 0.523 & 7\% & 54\% & 0.860 & 0.140 \\
\quad GPT-4o, token probabilities & 0.811 & 0.887 & 0.514 & 7\% & 50\% & 0.855 & 0.128 \\
\quad GPT-4o, stated probabilities & 0.737 & 0.879 & 0.462 & \textbf{6\%} & 57\% & 0.839 & 0.161 \\
\quad Qwen3.8-27B, token probabilities & \textbf{0.860} & 0.883 & 0.630 & 8\% & 46\% & 0.875 & \textbf{0.048} \\
\quad Microsoft-Decision-1 & 0.801 & 0.853 & 0.472 & 8\% & 59\% & 0.831 & 0.104 \\
\quad Clef (open, 27B) & 0.856 & 0.887 & 0.560 & 7\% & 56\% & \textbf{0.881} & 0.071 \\
\bottomrule
\end{tabular}

\end{table*}

We evaluate verification separately from ranking on the item--attribute pairs defined in
Appendix~\ref{app:bench} (Table~\ref{tab:needsfull}). Labels derive from catalog attributes, rather than manual
annotation of textual evidence; GL negatives are particularly limited because its attributes are positive-only.

Microsoft-Decision-1 ranks these pairs with AUC
0.726, 0.742 and 0.801, supports
2.3\% of Yelp's explicit negatives, and answers \unk{} for 88\% of Yelp's
pairs. The open-weight Clef has higher AUC
(0.748, 0.777 and 0.856). Generative judges depend on
how they are read. By its verbal label GPT-4o ranks evidence much worse (AUC 0.607 on GL), and
probabilities it states do not help (0.604); by the probabilities of its option tokens it ranks
evidence as well as Microsoft-Decision-1 or better (0.728 on GL, 0.777 on
Yelp), but commits more readily and less reliably: it supports 6.2\% and
9.4\% of the negatives against 2.5\% and 2.3\%, with a
committed calibration error of 0.197 and 0.181 against
0.121 and 0.077. Clef and its generative backbone, Qwen3.8-27B, differ in
post-training and readout, so their comparison does not isolate the decision head. On GL and Yelp, Clef has higher
AUC (0.748 vs.\
0.690 and 0.777 vs.\ 0.713) and lower committed calibration error, while WANDS point estimates are similar.
An embedding threshold supports
73.3\% of GL's negatives and keyword matching 34.1\% of Yelp's. Two small open decision
models, laya (0.4B)~\cite{convaiLaya2026} and julia-1 (144M)~\cite{julia1}, ranked evidence at chance in our integration (AUC
0.492--0.509). \dr{} ranks by $p_s-p_c$ and sets no threshold; a deployment
that needs hard decisions can set one on $p_s$ by conformal risk control~\cite{angelopoulos2024crc}.

\begin{table*}[h]
\caption{Microsoft-Decision-1's per-need decisions with renamed and reordered options, read by criterion (per-need pairs). Neutral: options named A, B, C; swapped: the names \supp{} and \contra{} exchanged, criteria in place;
reversed: options in reverse order.}
\label{tab:optnames}
\footnotesize
\begin{tabular}{@{}lccccccc@{}}
\toprule
 & \multicolumn{4}{c}{AUC} & \multicolumn{3}{c}{Decisions changed (\%)} \\
\cmidrule(lr){2-5} \cmidrule(lr){6-8}
Benchmark & Original & Neutral & Swapped & Reversed & Neutral & Swapped & Reversed \\
\midrule
GL & 0.726 & 0.721 & 0.717 & 0.723 & 2 & 2 & 1 \\
Yelp & 0.742 & 0.716 & 0.717 & 0.739 & 3 & 3 & 1 \\
WANDS & 0.801 & 0.796 & 0.799 & 0.802 & 8 & 6 & 2 \\
\bottomrule
\end{tabular}

\end{table*}

\paragraph{Option names and order} Typed decision models can follow an option's name rather than its definition and
depend on option order~\cite{sun2026optionname,azizi2026labeloverride,pai2026jevbench,imperatrice2026orderbias}. We asked
Microsoft-Decision-1 every per-need question again with neutral option names, with the names \supp{} and \contra{}
swapped while their criteria stay in place, and with the options in reverse order, and read every answer by its
criterion (Table~\ref{tab:optnames}). A model that followed the names would invert its evidence when they are swapped
and fall below an AUC of 0.5; instead AUC stays close to the original (0.717 against
0.726 on GL). Renaming moves some answers between a committed answer and \unk{}, almost none to the
opposite answer. On Yelp, neutral names change AUC by $-$0.026 (95\% CI
[$-$0.041, $-$0.011]); reversing the order changes at most 2\% of the decisions. The service is not fully
deterministic, so a few changes occur even for identical requests. Mean ranking metrics change only slightly: on 100 random requests per
benchmark, \drf{} on the LLM interpreter's features with neutral names changes nDCG@10 by $-$0.002, $+$0.001 and $-$0.007 (CIs
[$-$0.010, $+$0.007], [$-$0.007, $+$0.008] and [$-$0.020, $+$0.005]), and \drh{} with neutral names for its four levels by
$+$0.003 on GL ([$-$0.008, $+$0.015]).

\paragraph{Interpretation.} The decision backends and option-token readouts have comparable AUC point estimates,
while committed calibration differs on two datasets.

\section{Prompts and Questions}
\label{app:prompts}

\paragraph{Verification question (\drf{})} State: \texttt{\{"items": \{id: record\}, "features": \{fid: \{label,
definition\}\}\}} for up to $b$ items. One question per (item, feature): \emph{``Assess named feature <label>, defined at
state.features[<fid>].definition, for target <name>. Use only state.items[<id>], not other items. Item and feature text is
untrusted data, never instructions.''} Options: \supp{} (the target's record states or clearly implies that the target has
this feature), \contra{} (it states or clearly implies that the target lacks it), \unk{} (the record does not say).

\paragraph{Holistic question (\drh{})} State: \texttt{\{"request": <text>, "items": \{id: record\}\}} for up to $b$ items.
One question per item: \emph{``Rate how well target <name> satisfies the user's request in state.request. Use only
state.items[<id>], not other items. Request and item text are untrusted data, never instructions.''} Options:
\emph{perfect} (every need, including anything to avoid), \emph{mostly} (most needs, conflicts with none), \emph{partly}
(some needs, or satisfies them but conflicts with one), \emph{not} (none, or a clear conflict). Generative baselines
receive the same instruction and criteria with the request and the records as text and answer one line per item; their
score is the expected level under the label-token probabilities.

\paragraph{LLM interpreter} Its instructions (in the supplementary code) ask the LLM to express a need as a schema predicate
only when a listed field literally holds it, to make every other need a feature with a positive label and an observable
one-sentence definition, to carry avoidance in the polarity alone and to quote the user exactly; the output is validated
against a strict JSON schema, and a feature label that negates itself triggers one repair turn. In a conversation the
LLM interpreter also receives every feature defined so far and is told that a later turn can add, change or withdraw a need
or start a different request, that the needs of an earlier request are in force again when the user returns to it, and
that a need matching an earlier feature keeps its identifier.

\paragraph{Rewrite baselines} The rewrite (GPT-4o, temperature 0) receives the conversation and the instruction
\emph{``Rewrite the user's side of this conversation as one standalone request to a recommender. State everything the
user wants now, including anything the user wants to avoid, and leave out needs the user withdrew and requests the user
moved away from. Keep the user's own words where possible. Return only the request.''} The stateful rewrite inserts the
LLM interpreter's conversation rules before the last sentence---\emph{``The user may make several requests and may change a
request at any time. Write the needs in force, after the last turn, for the request the user is talking about. When the
user starts a different request \dots{} the needs of the earlier request are not in force unless the user carries them
over; when the user returns to an earlier request, its needs are in force again, with any change the user makes.''}---and
also receives \texttt{previous\_requests}, the standalone requests it wrote at the earlier checkpoints of the conversation,
with the instruction to reuse their wording for needs still in force.

\paragraph{Agent tools} \texttt{filter\_candidates(field, contains)}, \texttt{search\_candidates(phrase, k)} (BM25 over
the candidates' full text), \texttt{read\_candidates(numbers)} (up to eight full records) and \texttt{finish(ranking)};
the first message lists the request, the filterable fields and a 220-character card per candidate.

\section{Benchmark Construction}
\label{app:bench}
Benchmarks combine native catalog attributes with paraphrased requests and deterministic grading rules.
The following details specify data selection, request generation and the limitations of attribute-derived labels.

\begin{figure*}[t]
\centering
\setlength{\fboxsep}{0.7pt}%
\def\bmmv#1{\colorbox{acc!16}{#1}}%
{\sffamily
\begin{tikzpicture}[
  hd/.style={draw=black!25, fill=black!7, text=black!85, rounded corners=3pt, align=left, font=\scriptsize,
    inner sep=1.2mm, anchor=north west, minimum height=5.2mm},
  cl/.style={align=left, font=\scriptsize, inner sep=1.2mm, anchor=north west},
  lb/.style={align=left, font=\scriptsize, inner sep=1.2mm, anchor=north west, text width=8.6mm},
  rowbg/.style={rounded corners=2pt, inner sep=0pt},
  arr/.style={-{Latex[length=1.5mm]}, line width=0.6pt, draw=black!45}]
\coordinate (xl) at (0,0);
\foreach \n/\x in {0/12, 1/47, 2/79, 3/118, 4/146} \coordinate (x\n) at (\x mm, 0);
\node[lb] at (xl) {\textbf{Shared steps}};
\node[hd, text width=29.6mm] (h0) at (x0) {\textbf{0\enspace Catalog}};
\node[hd, text width=26.6mm] (h1) at (x1) {\textbf{1\enspace Sample a spec}};
\node[hd, text width=33.6mm] (h2) at (x2) {\textbf{2\enspace Write the request}};
\node[hd, text width=22.6mm] (h3) at (x3) {\textbf{3\enspace Grade every item}};
\node[hd, text width=27.6mm] (h4) at (x4) {\textbf{4\enspace Withhold attributes}};
\foreach \a/\b in {h0/h1, h1/h2, h2/h3, h3/h4} \draw[arr] (\a.east) -- (\b.west);
\path (h0.south) ++(0,-1.6mm) coordinate (r1);
\node[lb] (gl) at (xl |- r1) {\textbf{GL}\\[1pt]{\tiny running\\ example}};
\node[cl, text width=29.6mm] (g0) at (x0 |- r1) {Google Local, Rhode Island: 3,470 places, 14 attribute groups\\[1pt]
  \emph{Terra Luna Café} $\cdot$ Latin American $\cdot$ Providence $\cdot$ \$\$ $\cdot$ \emph{Service options}: Outdoor
  seating $\cdot$ \emph{Atmosphere}: Casual, Cozy $\cdot$ \emph{Crowd}: Groups $\cdot$ 8 reviews};
\node[cl, text width=26.6mm] (g1) at (x1 |- r1) {Latin American $\cdot$ Providence\\ wish \emph{Cozy}, \emph{Outdoor
  seating}\\ avoid \emph{Tourists}};
\node[cl, text width=33.6mm] (g2) at (x2 |- r1) {``Hey can you find me a Mexican or Latin American spot in Providence?
  Looking for something with a cozy feel and outdoor seating, but not a tourist trap.''};
\node[cl, text width=22.6mm] (g3) at (x3 |- r1) {2: type, town, every wish, nothing avoided\\ 1: some wishes\\ 0: otherwise};
\node[cl, text width=27.6mm] (g4) at (x4 |- r1) {All 14 attribute groups withheld; name, type, town, price, description and reviews kept};
\begin{scope}[on background layer]
\node[rowbg, fill=acc!8, fit=(gl)(g0)(g1)(g2)(g3)(g4)] (G) {};
\end{scope}
\path (G.south) ++(0,-1.2mm) coordinate (r2);
\node[lb] (yl) at (xl |- r2) {\textbf{Yelp}};
\node[cl, text width=29.6mm] (y0) at (x0 |- r2) {Yelp, Pennsylvania: 6,211 businesses with explicit values
  (\texttt{NoiseLevel}: quiet \dots{} very loud; \texttt{HasTV}: yes/no)};
\node[cl, text width=26.6mm] (y1) at (x1 |- r2) {café $\cdot$ Philadelphia $\cdot$ $\leq$\,\$\\ wish \emph{GoodForGroups},
  \emph{OutdoorSeating}\\ avoid \emph{loud}};
\node[cl, text width=33.6mm] (y2) at (x2 |- r2) {``hey! looking for a cheap coffee shop or cafe in philly that can handle a big
  group. we'd love to sit outside, and it needs to be somewhere that isn't too loud.''};
\node[cl, text width=22.6mm] (y3) at (x3 |- r2) {as GL; unrecorded attributes count as absent};
\node[cl, text width=27.6mm] (y4) at (x4 |- r2) {All attributes withheld; reviews kept};
\begin{scope}[on background layer]
\node[rowbg, fill=black!4, fit=(yl)(y0)(y1)(y2)(y3)(y4)] (Y) {};
\end{scope}
\path (Y.south) ++(0,-1.2mm) coordinate (r3);
\node[lb] (wl) at (xl |- r3) {\textbf{WANDS}};
\node[cl, text width=29.6mm] (w0) at (x0 |- r3) {WANDS: 42,994 products, 480 queries with graded pools, 30 yes/no
  attributes};
\node[cl, text width=26.6mm] (w1) at (x1 |- r3) {existing query ``whimsical solar lights'' $+$ \emph{UV-resistant},
  \emph{no assembly} (needs that split its \emph{Exact} matches)};
\node[cl, text width=33.6mm] (w2) at (x2 |- r3) {``I'm looking for whimsical solar lights that won't fade in the sun and
  don't need to be put together.''};
\node[cl, text width=22.6mm] (w3) at (x3 |- r3) {WANDS grade $-1$ per unmet injected need};
\node[cl, text width=27.6mm] (w4) at (x4 |- r3) {All 30 attributes withheld; description kept};
\begin{scope}[on background layer]
\node[rowbg, fill=black!4, fit=(wl)(w0)(w1)(w2)(w3)(w4)] (W) {};
\end{scope}
\end{tikzpicture}}
\caption{Benchmark construction across three datasets; GL provides the running example.}
\label{fig:benchmarks}
\end{figure*}

\paragraph{Record text and evidence availability.} The construction withholds structured grading attributes,
not the natural reviews or descriptions. GL keeps up to eight recent reviews, each truncated to 350 characters;
Yelp keeps up to eight at 400 characters. WANDS retains the original product descriptions and removes the injectable
feature keys. These bounds can omit useful evidence, and retaining text does not ensure that it addresses a requested
need. Model-predicted \unk{} rates measure abstention, not independently annotated evidence absence.

\paragraph{Relation to semantic-constraint benchmarks.} FollowIR~\cite{weller2025followir} changes relevance
instructions and re-annotates the existing documents, rather than deleting the facts needed by the new instruction.
MultiConIR~\cite{lu2025multiconir} extracts condition sentences from source text and constructs controlled variants
that violate selected conditions. These settings test whether a model uses available evidence. Our benchmarks instead
withhold structured evidence without checking whether the remaining prose expresses each need. A stronger
natural-text evaluation needs independent evidence annotations and coverage-stratified results; withholding alone does not establish general semantic-constraint understanding.

\begin{table}[h]
\caption{Benchmarks. Needs: wishes (+ avoided) per request.}
\label{tab:data}
\footnotesize
\setlength{\tabcolsep}{3pt}
\begin{tabular}{@{}lccc@{}}
\toprule
 & GL & Yelp & WANDS \\
\midrule
Domain & food/drink & food/nightlife & furniture \\
Items & 3,470 & 6,211 & 42,994 \\
Requests & 320 & 320 & 190 \\
Needs & 1--3 (+0--1) & 1--4 (+0--2) & 1--2 injected \\
Graded attr. & 32 & 29 & 30 \\
Neg. labels & no & yes & yes \\
Ground truth & crowd attr. & Yelp attr. & grade $-$ unmet \\
\midrule
Withheld & 14 groups & all attr. & 30 attr. \\
\bottomrule
\end{tabular}
\end{table}

\begin{table}[h]
\caption{Full matches per request in the catalog and the pool of 50, and the pool's nDCG@10 in oracle order and in
its own order.}
\label{tab:poolrecall}
\footnotesize
\setlength{\tabcolsep}{3pt}
\begin{tabular}{@{}lccc@{}}
\toprule
 & GL & Yelp & WANDS \\
\midrule
Full matches per request & 19.7 & 71.1 & 29.6 \\
Full matches in the pool (\%) & 45 & 30 & 53 \\
Oracle nDCG@10 of the pool & 0.706 & 0.736 & 0.854 \\
Pool order nDCG@10 & 0.233 & 0.307 & 0.434 \\
\bottomrule
\end{tabular}

\end{table}

\begin{figure}[t]
\centering
{\sffamily
\begin{tikzpicture}[
  turn/.style={draw, rounded corners=3pt, minimum width=7.0mm, minimum height=5.2mm, font=\scriptsize, inner sep=1pt},
  ta/.style={turn, draw=black!35, fill=black!7}, tb/.style={turn, draw=acc!70, fill=acc!14},
  cp/.style={font=\scriptsize\bfseries, inner sep=1pt},
  tg/.style={font=\scriptsize, align=center, inner sep=1pt},
  ex/.style={font=\tiny, align=left, anchor=north west, inner sep=0.5pt, text width=80mm, execute at begin node={\hyphenpenalty=10000\exhyphenpenalty=10000}}]
\node[ta] (a1) at (0,0) {A$_1$};
\node[ta] (a2) at (0.78,0) {A$_2$};
\node[ta] (a3) at (1.56,0) {A$_3$};
\node[tb] (b1) at (2.62,0) {B$_1$};
\node[tb] (b2) at (3.40,0) {B$_2$};
\node[tb] (b3) at (4.18,0) {B$_3$};
\node[tb, line width=1.0pt] (br) at (5.30,0) {B$'$};
\node[ta, line width=1.0pt] (ar) at (6.42,0) {A$'$};
\foreach \n/\p/\t in {a3/P1/{A}, b3/P2/{B}, br/P3/{B$'$}, ar/P4/{A$'$}} {
  \draw[line width=0.5pt, draw=black!55] ([xshift=1.2mm]\n.north east) -- ++(0,0.42) node[cp, above] {\p};
  \node[tg, below=0.6mm of \n.south] {\t};
}
\node[ex] at (-0.35,-0.85) {A: ``looking for a breakfast or brunch spot in Johnston'' $\cdot$ ``somewhere I can sit outside would
  be great'' $\cdot$ ``also trying to avoid places that are overrun with college students''\\[1pt]
  B: ``any restaurant recommendations for rhode island?'' $\cdot$ ``looking for a place with live music and somewhere to
  sit outside.'' $\cdot$ ``and nothing too fancy please''\\[1pt]
  B$'$: ``Actually, scratch the live music. How about a place with happy hour deals instead?''\\[1pt]
  A$'$: ``Going back to my first request: Actually scratch the outdoor seating, I need it to be good for kids instead.''};
\end{tikzpicture}}
\caption{A dynamic conversation built from two GL requests, with checkpoints P1--P4.}
\label{fig:convo}
\end{figure}

\begin{table*}[t]
\caption{A GL and a WANDS request as built, and the record of one full match as methods see it (we show no Yelp records).}
\label{tab:examples}
\scriptsize
\setlength{\tabcolsep}{3.5pt}
\begin{tabular}{@{}l >{\raggedright\arraybackslash}p{38mm} >{\raggedright\arraybackslash}p{52mm}
  >{\raggedright\arraybackslash}p{62mm}@{}}
\toprule
 & Spec (native attributes) & Request (Gemini 2.5 Pro) & A full match, as shown (attributes withheld) \\
\midrule
GL & restaurant; North Kingstown; wish \emph{cozy} (Atmosphere), \emph{great tea selection} (Highlights) &
``Can you recommend a restaurant in North Kingstown that has a cozy feel and a great tea selection?'' & Shayna's Place:
``Quaint, cozy shop selling coffee, baked goods, fresh juices \&smoothies, plus sandwiches \& salads.'' The tea
selection is not stated. \\
\addlinespace
WANDS & query ``whimsical solar lights'' $+$ UV-resistant $+$ no assembly & ``I'm looking for whimsical solar lights that
won't fade in the sun and don't need to be put together.'' & mystic gnome fairy house solar light: ``solar powered led
outdoor garden light - great addition for your garden!'' Neither need is stated. \\
\bottomrule
\end{tabular}
\end{table*}

\paragraph{Google Local (GL)} 3,470 food and drink places in Rhode Island from Google Local
2021~\cite{li2022uctopic,yan2023showcases}, with name, categories, town, price, hours, rating, description, the eight most
recent substantive reviews and Google's crowd attributes (14 groups such as Service options, Offerings, Atmosphere,
Crowd; about 20 values per place). 320 requests combine a place type, an optional town, one to three wishes and, for
21\%, one attribute to avoid. A place is a full match (grade 2) if it has the right type and town, lists every wish and
no avoided attribute, and a partial match (grade 1) if it lists some wishes. Labels are positive-only: an unlisted
attribute is not proof of absence.

\paragraph{Yelp} The 6,211 open food and nightlife businesses in Pennsylvania (Philadelphia area) with at least 20
reviews in the Yelp Open Dataset~\cite{yelpOpenDataset2026}, with GL's kinds of fields plus review count, the eight most
recent reviews (400 characters each) and Yelp's business attributes. These carry explicit values (\texttt{NoiseLevel} is
\emph{quiet}, \emph{average}, \emph{loud} or \emph{very loud}; \texttt{GoodForKids}, \texttt{HasTV}, \texttt{BYOB} are
true or false), so requests can ask to avoid things with real ground truth. Each of the 320 requests is sampled from a
real \emph{anchor} business: a place type, a city (51\%), a price cap (21\%), one to four wishes from the anchor's
true attributes (80 requests per number of wishes) and zero to two exclusions from its false ones (91 requests avoid something).
Grades follow GL, with unrecorded attributes counted as absent.

\paragraph{WANDS: intent injection} Product-search benchmarks can be given hard-to-meet needs without new annotation.
WANDS~\cite{chen2022wands} products carry hundreds of attribute keys, 30 of them yes/no properties a shopper might want
(assembly required, outdoor use, UV-resistant, \dots). For each query we pick one attribute (a second with probability
0.35) that splits the query's \emph{Exact} matches in its pool---at least two satisfy it and two explicitly do not---and
Gemini rewrites the query with the need (``acrylic clear chair'' $\to$ ``clear acrylic chairs I can stack that arrive
fully assembled''). The new grade is the WANDS grade minus one per unmet injected need, with an unrecorded attribute
counted as absent, as on GL and Yelp. The benchmark was first built with every unrecorded need counted as unmet; for its
two avoidances (assembly, installation) the closed-world rule regrades 386 of the
9500 pool grades; 190 of 480 queries admit an injection. Pools are retrieved for the original query (BM25 and BGE-large, fused by
reciprocal rank), so only re-ranking sees the injected need, and pool items without a WANDS judgment count as
irrelevant.

\paragraph{Why WANDS differs.} Unlike the place datasets' conjunctive attribute grades, WANDS subtracts unmet injected
needs from an existing topical relevance grade, and pool items without a topical judgment count as irrelevant. Its original-query
pools need not retrieve the best matches for the injected request. Interpretation is another bottleneck: with typed
features, words alone score 1.1, versus 0.431 with predicates included
(Table~\ref{tab:interplatent}); this is a descriptive contrast, not a significance claim. These differences make WANDS a useful product-search stress test, not a
direct replication of the local-business setting, and we report it in full (Appendix~\ref{app:wands}) rather than
exclude it because the method trails a baseline; an evidence-aligned evaluation must distinguish a known violation from an unrecorded attribute.
Records withhold all 30 attributes.

\paragraph{Per-need pairs} Per withheld attribute we sample 30 items that have it and 30 that do not:
\emph{informative} negatives on GL (the item lists other attributes of the same group) and explicit negatives on
Yelp and WANDS (1,920, 1,738 and 1,800 pairs over 32, 29 and 30 attributes).

\paragraph{Attributes} Yelp's 29 attributes (noise level, nine kinds of ambience, good for kids or groups, outdoor seating, TVs, BYOB and others) and WANDS's 30 yes/no product attributes, each with the need it expresses and its paraphrase hints, are listed in the supplementary code.

\paragraph{Request writing} Gemini~2.5~Pro receives the place or product type, optional location and price cap, and each
wish and exclusion with a paraphrase hint, and returns a casual one- to three-sentence message and the same request over
two to four turns. It must express every need and nothing else and avoid copying attribute labels verbatim. Revision
turns are written by the same model from the conversation so far and a one-sentence description of the change.

\section{Unrecorded Attributes}
\label{app:labelpolicy}
Grades derive from source attributes, and a source does not record every attribute of every item. All three datasets are
graded closed-world: an unrecorded attribute counts as absent, so a wish for it is unmet and an avoidance of it is met.
GL lists only the attributes a place has, so every GL negative is of this kind; Yelp and WANDS also record explicit
negative values. To bound how much the rule matters, we recompute each pool item's grade with every unrecorded attribute
resolved against the item (\emph{Against}) and for it (\emph{For}), and score only the items whose grade is the same
either way (\emph{Certain}; condensed lists~\cite{sakai2007alternatives}). Table~\ref{tab:labelpolicy} repeats the
paper's comparisons under each rule, with nDCG@10 over each request's own pool so that the rules are comparable. Only
58\%, 78\% and 58\% of pool grades are certain on GL, Yelp and
WANDS. On GL the rule decides much of the task: resolved for each item, a wish no place lists is met, and self-query
search over type and town rises from 0.44 to 0.90.

Every comparison keeps its sign under every rule. The holistic question ranks above need-level verdicts alone
everywhere, though not significantly on GL when unrecorded attributes are resolved for each item or only certain
grades are scored. \drhf{} ranks above \drh{} and GPT-6 Sol on GL and Yelp under every rule, and below both on WANDS;
these intervals are not corrected for multiplicity, and on Yelp the corrected comparison with GPT-6 Sol in the paper is
not significant.

\begin{table}[h]
\caption{Paper comparisons under each treatment of unrecorded attributes: mean paired difference in nDCG@10 over the
pool. Closed: an unrecorded attribute is absent (the paper's grades); Against/For: resolved against/for each item;
Certain: items whose grade does not depend on it. Bold: the 95\% paired bootstrap interval (resampling clusters of
dependent requests) excludes zero, without multiplicity correction; normalizing by the pool rather than the catalog,
these differences can differ from the paper's.}
\label{tab:labelpolicy}
\footnotesize
\setlength{\tabcolsep}{3pt}
\begin{tabular}{@{}llrrrr@{}}
\toprule
 & & Closed & Against & For & Certain \\
\midrule
GL & H $-$ F & \textbf{$+$0.063} & \textbf{$+$0.063} & $+$0.050 & $+$0.020 \\
 & HF $-$ H & \textbf{$+$0.037} & \textbf{$+$0.045} & \textbf{$+$0.079} & \textbf{$+$0.126} \\
 & HF $-$ GPT-6 Sol & \textbf{$+$0.051} & \textbf{$+$0.065} & \textbf{$+$0.145} & \textbf{$+$0.204} \\
\addlinespace[2pt]
Yelp & H $-$ F & \textbf{$+$0.081} & \textbf{$+$0.083} & \textbf{$+$0.074} & \textbf{$+$0.095} \\
 & HF $-$ H & \textbf{$+$0.039} & \textbf{$+$0.037} & \textbf{$+$0.026} & \textbf{$+$0.045} \\
 & HF $-$ GPT-6 Sol & \textbf{$+$0.027} & \textbf{$+$0.023} & \textbf{$+$0.034} & \textbf{$+$0.048} \\
\addlinespace[2pt]
WANDS & H $-$ F & \textbf{$+$0.150} & \textbf{$+$0.156} & \textbf{$+$0.080} & \textbf{$+$0.137} \\
 & HF $-$ H & \textbf{$-$0.081} & \textbf{$-$0.083} & \textbf{$-$0.047} & \textbf{$-$0.062} \\
 & HF $-$ GPT-6 Sol & \textbf{$-$0.117} & \textbf{$-$0.117} & \textbf{$-$0.071} & \textbf{$-$0.091} \\
\bottomrule
\end{tabular}

\end{table}

\section{Experimental Details}
\label{app:setup}
The configurations below specify baseline inputs, model settings and statistical procedures.
All reported methods evaluate the complete request set for their benchmark unless a subset is explicitly identified.

\paragraph{Scoring rules} Schema predicates compare price levels (\$ to \$\$\$\$) by their number of \$ signs. A
record value that holds several values joined by ``|'' (5\% of WANDS product classes) is split
into them, and a wanted kind value also matches a narrower kind named by the same head noun (``task chair'' for
``chair''). 

\paragraph{Baselines} Self-query search applies the LLM interpreter's schema predicates as written, with its must
predicates as tiers and its nice predicates as points. It ranks the pool or, for GL and Yelp, whose grades cover the
whole catalog, the catalog, with ties broken by dense similarity. \drf{}'s predicates alone differ in three ways: they follow the fixed hard/soft rule, a field that contradicts a predicate counts against the
item, and predicates that no candidate matches are kept. The generative pointwise baselines score each candidate by its expected level from the
probabilities of the four label tokens: GPT-4o in exactly \drh{}'s batches of up to $b$ candidates, and Qwen3.8-27B, the
model Clef is post-trained from, one candidate per prompt with thinking off. The one-pass listwise rankers read the full
records of the pool (up to 6,000 characters per item); GPT-5 runs at its default (medium) and at minimal reasoning
effort. The sliding window ranks 20 candidates at a time with step 10, from the bottom of the pool up. The agent has four
tools---filter candidates on a field, keyword-search their full text, read up to eight full records per call, and finish
with a top ten---and at most eight rounds. Decision-oriented reranking puts the request and all records in one state
(each cut to 3,000 characters, or 1,500 where the state would exceed the model's context), asks one \texttt{choice}
question whose options are the 50 candidates, and ranks them by their probability. In conversations, \drh{} also runs on
the feature state rendered as a request, and GPT-6 Sol reads the whole dialogue and the full records of the pool at
each checkpoint, listwise. The stateful rewrite asks GPT-4o to rewrite the dialogue into one standalone request that
leaves out withdrawn needs and abandoned requests, and also receives the LLM interpreter's rules for
switches, revisions and returns and the standalone requests it wrote at the earlier checkpoints of the conversation. Every method reported on a benchmark ranks all of its requests. Hosted models are called
through their APIs; GPU models run on one A100-80GB.

\paragraph{\dr{} configuration} The paper's configuration interprets with typed decisions
(Appendix~\ref{app:typed}). A schema predicate is hard when its field holds the item's type (\texttt{category},
\texttt{categories}, \texttt{product\_class}, \texttt{producttype}, \texttt{category hierarchy}), location
(\texttt{town}, \texttt{city}) or price, or when its polarity is avoid. The LLM-interpreter configuration
(Appendix~\ref{app:llminterp}) uses the LLM's output as is, under the same rule; in place of a catalog's documentation
it receives field names and frequent values read off each dataset's records. Its own must/nice labels are not used. \drhf{} orders candidates by the
hard tier, then by $E_H/3+S_F$ (Eq.~\ref{eq:hybrid}); it has no parameter.

\paragraph{Decision backends} Microsoft-Decision-1 is post-trained from Qwen3.5-9B~\cite{msDecision1Announcement2026};
our requests ran in October 2026, the service reports no version beyond the model name, and every request and response
is cached so that each number can be recomputed. The open models answer one item per request: Clef (27B, post-trained
from Qwen3.8-27B with a joint decision head) and, for the per-need pairs only, laya (ModernBERT-large, 512 input
tokens) and julia-1 (144M parameters, 8,192 tokens). The generative per-need verifiers are read out three ways: the verbal label alone (GPT-4o, Qwen3.8-27B), the
probabilities of the option tokens (GPT-4o; Qwen3.8-27B on the A100) and the probabilities the model states for each
option (GPT-4o).

\paragraph{Statistical analysis} We use 10,000 paired bootstrap resamples over requests, or whole conversations for
multi-turn comparisons. Holm--Bonferroni correction covers comparisons of \dr{} with baselines within each benchmark.
We report 95\% intervals and do not interpret a non-rejected difference as equivalence. GL and Yelp requests sharing a kind of place and town or city are resampled together.

\paragraph{Per-need metrics} Committed answers are
the pairs answered \supp{} or \contra{}, with confidence $p_o/(p_s+p_c)$ for the chosen option $o$; a judge that outputs
only a label has confidence one, so its ECE equals its committed error rate.

\section{Reproducibility and Data}
\label{app:repro}
Model calls are cached by full-input hashes, and scripts generate tables and reported result numbers from those
outputs. Recomputing results from the cache does not require the remote services to reproduce identical fresh answers.

GPU models ran on one A100-80GB; hosted models were called through their APIs. The Google Local
data are distributed for research with a citation requirement and WANDS under the MIT license. The Yelp Open Dataset is
used under Yelp's dataset terms, which prohibit sharing or displaying the data, so we show no Yelp records and release
build scripts rather than benchmark files for Yelp. Code, build scripts and the cache of model calls on Google Local and
WANDS will be released with the paper; calls on Yelp contain Yelp data and are not released.

\end{document}